\documentclass{article}
\usepackage{iclr2027_conference,times}

\usepackage[utf8]{inputenc}
\usepackage[T1]{fontenc}
\usepackage{amsmath,amssymb,amsthm}
\usepackage{mathtools}
\usepackage{bm}
\usepackage{booktabs}
\usepackage{multirow}
\usepackage{graphicx}
\usepackage{xcolor}
\usepackage{colortbl}
\usepackage{tikz}
\usetikzlibrary{arrows.meta,positioning,calc,fit,backgrounds,shapes.misc,decorations.pathreplacing}
\usepackage{enumitem}
\usepackage{microtype}
\usepackage{xspace}
\usepackage{subcaption}
\usepackage{float}
\usepackage{algorithm}
\usepackage{algpseudocode}
\usepackage{pifont}
\usepackage{hyperref}
\usepackage{url}
\hypersetup{
  colorlinks=true,linkcolor={blue!60!black},citecolor={blue!60!black},urlcolor={blue!60!black},
  pdftitle={Toaster: Nonlinear Memory from Cascaded Linear Memories and Target Propagation},
  pdfauthor={Zeman Li, Ali Behrouz, Yuan Deng, Praneeth Kacham, Rami Al-Rfou, Meisam Razaviyayn, Vahab Mirrokni},
}

\newcommand{\model}{\textsc{Toaster}\xspace}

\newcommand{\R}{\mathbb{R}}
\newcommand{\I}{\mathbf{I}}
\DeclareMathOperator{\Diag}{Diag}
\DeclareMathOperator{\diag}{diag}
\DeclareMathOperator{\softsign}{softsign}

\newcommand{\T}{\mathsf{T}}                 %
\newcommand{\had}{\odot}

\newcommand{\vq}{\bm{q}}
\newcommand{\vk}{\bm{k}}
\newcommand{\vv}{\bm{v}}
\newcommand{\vo}{\bm{o}}
\newcommand{\vz}{\bm{z}}
\newcommand{\vh}{\bm{h}}
\newcommand{\ve}{\bm{e}}
\newcommand{\vw}{\bm{w}}
\newcommand{\vu}{\bm{u}}
\newcommand{\vx}{\bm{x}}
\newcommand{\valpha}{\bm{\alpha}}
\newcommand{\vbeta}{\bm{\beta}}

\newcommand{\mS}{\mathbf{S}}
\newcommand{\mK}{\mathbf{K}}
\newcommand{\mQ}{\mathbf{Q}}
\newcommand{\mV}{\mathbf{V}}
\newcommand{\mO}{\mathbf{O}}
\newcommand{\mZ}{\mathbf{Z}}
\newcommand{\mH}{\mathbf{H}}
\newcommand{\mE}{\mathbf{E}}
\newcommand{\mW}{\mathbf{W}}
\newcommand{\mU}{\mathbf{U}}
\newcommand{\mT}{\mathbf{T}}
\newcommand{\mM}{\mathbf{M}}
\newcommand{\mP}{\mathbf{P}}

\newcommand{\Sone}{\mS^{(1)}}
\newcommand{\Stwo}{\mS^{(2)}}
\newcommand{\dk}{d_k}
\newcommand{\dv}{d_v}
\newcommand{\dhid}{d_h}
\newcommand{\vhat}{\hat{\vv}}
\newcommand{\vtil}{\tilde{\vv}}
\newcommand{\Vtil}{\widetilde{\mV}}
\newcommand{\Norm}{\mathcal{N}}
\newcommand{\Loss}{\mathcal{L}}
\newcommand{\Mem}{\mathcal{M}}
\newcommand{\kernel}{\textsc{LinMem}}

\newcommand{\ch}[2]{#1_{[#2]}}

\definecolor{bestcolor}{HTML}{DCE9F5}
\definecolor{secondcolor}{HTML}{F1F6FB}
\newcommand{\best}[1]{\cellcolor{bestcolor}\textbf{#1}}
\newcommand{\second}[1]{\cellcolor{secondcolor}#1}
\newcommand{\yes}{\textcolor[HTML]{2E7D32}{\ding{51}}}
\newcommand{\no}{\textcolor[HTML]{C62828}{\ding{55}}}

\newcommand{\secref}[1]{Section~\ref{#1}}
\newcommand{\appref}[1]{Appendix~\ref{#1}}
\newcommand{\figref}[1]{Figure~\ref{#1}}
\newcommand{\tabref}[1]{Table~\ref{#1}}
\newcommand{\eqnref}[1]{Eq.~\eqref{#1}}
\renewcommand{\algref}[1]{Algorithm~\ref{#1}}

\newtheorem{proposition}{Proposition}
\newtheorem{remark}[proposition]{Remark}

\definecolor{cAttn}{HTML}{6B7280}
\definecolor{cKDA}{HTML}{D97706}
\definecolor{cToaster}{HTML}{1D4ED8}
\definecolor{cSlot}{HTML}{DBEAFE}
\definecolor{cSlotBorder}{HTML}{1E40AF}
\definecolor{cAct}{HTML}{FDE68A}
\definecolor{cActBorder}{HTML}{B45309}
\definecolor{cTarget}{HTML}{FEE2E2}
\definecolor{cTargetBorder}{HTML}{B91C1C}
\definecolor{cGray}{HTML}{4B5563}

\newcommand{\ladderlossrows}{%
\addlinespace[2pt]\multicolumn{9}{@{}l}{\textit{C4 (in-domain)}} \\
\quad 145M & 2.7678 & 2.7725 & \best{2.7530} & 15.92 & 16.00 & \best{15.69} & $-$0.0195 & $-$0.0148 \\
\quad 300M & 2.6039 & 2.6082 & \best{2.5955} & 13.52 & 13.57 & \best{13.40} & $-$0.0127 & $-$0.0084 \\
\quad 550M & 2.4799 & 2.4820 & \best{2.4726} & 11.94 & 11.97 & \best{11.85} & $-$0.0094 & $-$0.0073 \\
\quad 850M & 2.3774 & 2.3822 & \best{2.3755} & 10.78 & 10.83 & \best{10.76} & $-$0.0067 & $-$0.0019 \\
\quad 1.2B & 2.2982 & 2.3027 & \best{2.2968} & 9.96 & 10.00 & \best{9.94} & $-$0.0059 & $-$0.0014 \\
\addlinespace[2pt]\multicolumn{9}{@{}l}{\textit{PG19}} \\
\quad 145M & 3.1094 & 3.1198 & \best{3.1005} & 22.41 & 22.64 & \best{22.21} & $-$0.0193 & $-$0.0089 \\
\quad 300M & 2.9299 & 2.9432 & \best{2.9275} & 18.73 & 18.98 & \best{18.68} & $-$0.0157 & $-$0.0024 \\
\quad 550M & 2.7973 & 2.8070 & \best{2.7954} & 16.40 & 16.56 & \best{16.37} & $-$0.0116 & $-$0.0019 \\
\quad 850M & \best{2.6753} & 2.6855 & 2.6792 & \best{14.52} & 14.66 & 14.57 & $-$0.0063 & $+$0.0039 \\
\quad 1.2B & \best{2.5793} & 2.5933 & 2.5869 & \best{13.19} & 13.37 & 13.29 & $-$0.0064 & $+$0.0076 \\}
\newcommand{\ladderfitrows}{%
C4 (in-domain) & Attention & 0.047 & 0.9997 & 1 & 1 & 1 & 1 & 1 \\
 & KDA & 0.047 & 0.9996 & 0.96 & 0.97 & 0.98 & 0.96 & 0.96 \\
 & \model & 0.045 & 0.9998 & 1.12 & 1.07 & 1.07 & 1.02 & 1.01 \\
\addlinespace[2pt]
PG19 & Attention & 0.047 & 0.9997 & 1 & 1 & 1 & 1 & 1 \\
 & KDA & 0.046 & 0.9998 & 0.93 & 0.91 & 0.93 & 0.92 & 0.89 \\
 & \model & 0.045 & 0.9999 & 1.06 & 1.02 & 1.01 & 0.97 & 0.94 \\}

\newcommand{\ladderpplwiderows}{%
Attention & 15.92 & 13.52 & 11.94 & 10.78 & 9.96 & 22.41 & 18.73 & 16.40 & \best{14.52} & \best{13.19} \\
KDA & 16.00 & 13.57 & 11.97 & 10.83 & 10.00 & 22.64 & 18.98 & 16.56 & 14.66 & 13.37 \\
\model & \best{15.69} & \best{13.40} & \best{11.85} & \best{10.76} & \best{9.94} & \best{22.21} & \best{18.68} & \best{16.37} & 14.57 & 13.29 \\}

\newcommand{\downstreamfullrows}{%
\addlinespace[2pt]\multicolumn{9}{@{}l}{\textit{Science QA}} \\
\quad ARC-Challenge & 30.64 & \second{31.16} & 30.84 & \best{32.62} & \best{33.39} & 32.02 & 29.36 & \second{32.19} \\
\quad ARC-Easy & 60.68 & \second{61.06} & 56.07 & \best{61.55} & \second{61.01} & 60.25 & 57.72 & \best{62.07} \\
\quad\textit{average} & 45.66 & \second{46.11} & 43.45 & \best{47.08} & \best{47.20} & 46.14 & 43.54 & \second{47.13} \\
\addlinespace[2pt]\multicolumn{9}{@{}l}{\textit{Commonsense reasoning}} \\
\quad COPA & 67.00 & 70.00 & \best{73.00} & \second{72.00} & 70.00 & 70.00 & \second{73.00} & \best{75.00} \\
\quad HellaSwag & \best{62.03} & \second{61.33} & 58.47 & 61.23 & \second{61.97} & \best{62.67} & 60.90 & 61.73 \\
\quad OpenBookQA & \second{38.60} & 38.40 & \best{39.20} & 38.40 & \best{40.80} & 37.80 & \second{39.40} & 38.00 \\
\quad PIQA & 75.14 & \second{75.19} & 73.72 & \best{76.16} & \second{75.63} & 74.97 & 74.97 & \best{75.92} \\
\quad SWAG & 61.20 & \best{62.47} & 61.20 & \second{62.17} & 61.63 & \second{62.43} & 61.67 & \best{62.93} \\
\quad\textit{average} & 60.79 & \second{61.48} & 61.12 & \best{61.99} & \second{62.01} & 61.57 & 61.99 & \best{62.72} \\
\addlinespace[2pt]\multicolumn{9}{@{}l}{\textit{Reading comprehension}} \\
\quad MultiRC & \second{57.16} & 57.12 & 57.05 & \best{57.20} & 57.59 & 57.18 & \second{57.63} & \best{58.16} \\
\quad\textit{average} & \second{57.16} & 57.12 & 57.05 & \best{57.20} & 57.59 & 57.18 & \second{57.63} & \best{58.16} \\
\addlinespace[2pt]\multicolumn{9}{@{}l}{\textit{Mathematical reasoning}} \\
\quad MathQA & \second{21.70} & 21.68 & 21.65 & \best{21.72} & \second{21.70} & 21.59 & 21.63 & \best{21.79} \\
\quad\textit{average} & \second{21.70} & 21.68 & 21.65 & \best{21.72} & \second{21.70} & 21.59 & 21.63 & \best{21.79} \\
\addlinespace[2pt]\multicolumn{9}{@{}l}{\textit{NLI \& entailment}} \\
\quad CB & \second{50.00} & \second{50.00} & \second{50.00} & \best{53.57} & \second{51.79} & 50.00 & 50.00 & \best{53.57} \\
\quad MNLI & 31.93 & \second{31.99} & 31.87 & \best{33.73} & \best{32.98} & \second{32.71} & 31.66 & 31.82 \\
\quad RTE & \second{53.79} & 47.29 & 48.38 & \best{54.51} & \second{53.43} & 50.35 & 50.18 & \best{55.12} \\
\quad\textit{average} & \second{45.24} & 43.09 & 43.42 & \best{47.27} & \second{46.07} & 44.35 & 43.95 & \best{46.84} \\
\addlinespace[2pt]\multicolumn{9}{@{}l}{\textit{Linguistic understanding}} \\
\quad SST-2 & 49.08 & 49.08 & 49.08 & 49.08 & 49.08 & 49.08 & 49.08 & 49.08 \\
\quad\textit{average} & 49.08 & 49.08 & 49.08 & 49.08 & 49.08 & 49.08 & 49.08 & 49.08 \\
\midrule
\textbf{Average over 13 tasks} & \second{50.69} & 50.52 & 50.04 & \best{51.84} & \second{51.62} & 50.85 & 50.55 & \best{52.11} \\}

\newcommand{\downstreammaingroups}{& \multicolumn{2}{c|}{Perplexity $\downarrow$} & \multicolumn{14}{c}{Downstream accuracy (\%) $\uparrow$} \\}
\newcommand{\downstreammainheader}{Model & C4 & PG19 & ARC-c & ARC-e & COPA & Hella. & OBQA & PIQA & SWAG & MultiRC & MathQA & CB & MNLI & RTE & SST-2 & Avg. \\}
\newcommand{\downstreammainrows}{%
\addlinespace[1pt]\multicolumn{17}{@{}l}{\textit{All layers replaced}} \\
Attention & \second{9.96} & \best{13.19} & 30.64 & 60.68 & 67.00 & \best{62.03} & \second{38.60} & 75.14 & 61.20 & \second{57.16} & \second{21.70} & \second{50.00} & 31.93 & \second{53.79} & 49.08 & \second{50.69} \\
KDA & 10.00 & 13.37 & \second{31.16} & \second{61.06} & 70.00 & \second{61.33} & 38.40 & \second{75.19} & \best{62.47} & 57.12 & 21.68 & \second{50.00} & \second{31.99} & 47.29 & 49.08 & 50.52 \\
Titans & 10.09 & 13.48 & 30.84 & 56.07 & \best{73.00} & 58.47 & \best{39.20} & 73.72 & 61.20 & 57.05 & 21.65 & \second{50.00} & 31.87 & 48.38 & 49.08 & 50.04 \\
\model & \best{9.94} & \second{13.29} & \best{32.62} & \best{61.55} & \second{72.00} & 61.23 & 38.40 & \best{76.16} & \second{62.17} & \best{57.20} & \best{21.72} & \best{53.57} & \best{33.73} & \best{54.51} & 49.08 & \best{51.84} \\
\midrule
\addlinespace[1pt]\multicolumn{17}{@{}l}{\textit{Hybrid (3 mixer : 1 attention)}} \\
SWA & 9.91 & 13.19 & \best{33.39} & \second{61.01} & 70.00 & \second{61.97} & \best{40.80} & \second{75.63} & 61.63 & 57.59 & \second{21.70} & \second{51.79} & \best{32.98} & \second{53.43} & 49.08 & \second{51.62} \\
KDA & \second{9.82} & \second{12.99} & 32.02 & 60.25 & 70.00 & \best{62.67} & 37.80 & 74.97 & \second{62.43} & 57.18 & 21.59 & 50.00 & \second{32.71} & 50.35 & 49.08 & 50.85 \\
Titans & 9.90 & 13.08 & 29.36 & 57.72 & \second{73.00} & 60.90 & \second{39.40} & 74.97 & 61.67 & \second{57.63} & 21.63 & 50.00 & 31.66 & 50.18 & 49.08 & 50.55 \\
\model & \best{9.78} & \best{12.75} & \second{32.19} & \best{62.07} & \best{75.00} & 61.73 & 38.00 & \best{75.92} & \best{62.93} & \best{58.16} & \best{21.79} & \best{53.57} & 31.82 & \best{55.12} & 49.08 & \best{52.11} \\}

\title{\model: Nonlinear Memory from Cascaded Linear Memories and Target Propagation}

\author{Zeman Li$^{1,2}$, Ali Behrouz$^{1}$, Yuan Deng$^{1}$, Praneeth Kacham$^{1}$, \\
  \textbf{Rami Al-Rfou$^{1}$, Meisam Razaviyayn$^{1,2}$ \& Vahab Mirrokni$^{1}$} \\
  $^{1}$Google Research \quad $^{2}$University of Southern California \\
  \texttt{\{zemanli,alibehrouz,dengyuan,pkacham\}@google.com} \\
  \texttt{\{rmyeid,razaviyayn,mirrokni\}@google.com}
}

\iclrfinalcopy
\makeatletter
\g@addto@macro\@maketitle{\lhead{Preprint}}
\makeatother

\begin{document}

\maketitle

\begin{abstract}
Recurrent memories let language models process long sequences with a fixed-size state.
Their design must balance the richness of the stored representation with the cost of updating it.
Linear memories such as DeltaNet and KDA store one matrix and take one delta-rule step per token.
The product of a chunk of such steps has an exact WY form, so they train fast.
On the other hand, deep memories such as TTT-MLP and Titans store a small MLP, which can hold maps that no matrix can.
However, their update has no WY form.
\model bridges the two by building on a known fact: a gradient step on a linear layer is a delta-rule write.
It builds a two-layer nonlinear memory from two linear memories with an activation between them.
We use \emph{target propagation} to give the hidden memory its values by projecting the output error back through the second memory. Given a chunk's cross-layer signals, approximated once per chunk, each layer can then run on an unmodified delta-rule kernel.
Our default runs this kernel on the output layer and updates the hidden layer once per chunk.
Running it on both layers gives a better model in our ablation. At matched backbones and training-token budgets, \model has lower loss than KDA at all five tested sizes (0.2B--1.6B non-embedding parameters) on both C4 and PG19.
Its C4 perplexity also matches or slightly improves on the attention baseline at every size.
At 1.2B it is also below Titans. In a 3:1 hybrid with global attention (one attention layer after every three mixer layers), the \model hybrid has the lowest perplexity of the four hybrids tested (KDA, Titans, sliding-window attention, and \model), and it is 0.18 (C4) and 0.44 (PG19) below the attention baseline.
At the largest size, \model's mean three-shot accuracy on 13 tasks is $51.84\%$ ($57.7\%$ on the seven not near chance), compared with $50.52\%$ ($57.1\%$) for KDA and $50.04\%$ ($56.1\%$) for Titans. Among the four hybrids, the \model hybrid again has the highest mean ($52.11\%$).
In length extrapolation, \model's loss stays below KDA's up to 20K tokens ($5\times$ the training length). On the 145M backbone, at a hidden width matched to Titans' memory, \model takes at most $1.33\times$ KDA's training-step time and is $2.8$--$5.5\times$ faster than the tested Titans configurations at 32K context.

\end{abstract}

\section{Introduction}
\label{sec:intro}

Softmax attention~\citep{vaswani2017attention} has revolutionized modern deep learning.
However, its key--value cache grows with the context, and a large body of work now replaces
some or all attention layers with a \emph{recurrent memory}: a fixed-size state that is written to and read from as the
sequence is processed~\citep{behrouz2026titans, katharopoulos2020transformers, gu2024mamba, yang2024gated,
yang2025gated}. Two questions separate one
memory from another: how much of the past it can pack into a given number of state
parameters, and how cheaply it can be trained.

\paragraph{Linear memory trains fast but stores a linear map.}
The memories that have been used in large open-source production models (e.g., Gated DeltaNet in
Qwen3-Next~\citep{qwen2025qwen3next}, Kimi Delta Attention (KDA) in Kimi Linear and the
2.8T-parameter Kimi K3~\citep{team2025kimi, kimik3}) are all \emph{linear}: the state is a
single matrix mapping keys to values, and each token applies one rank-one correction, a
generalized Householder step~\citep{schlag2021linear, yang2024parallelizing}. Products of
such factors admit the compact \emph{WY representation}~\citep{bischof1987wy}, which turns
a whole chunk of sequential updates into a few dense matrix multiplications while remaining
\emph{exact}; this is why these models train at throughput competitive with attention at
long context~\citep{yang2024parallelizing, yang2024fla}. However, their memory is restricted to linear mappings, which limits its expressivity.

\paragraph{Nonlinear memory stores more but its gradient update has no WY form.}
A second line of work, rooted in test-time memorization~\citep{sun2024learning,
behrouz2026titans, behrouz2025miras, behrouz2025atlas}, replaces the matrix with a small
neural network whose weights are updated online by gradient descent on a reconstruction
loss. Such a memory can store associations that no linear map can, but the joint gradient update
of its weights passes through the nonlinearity and is not affine in the state: no WY form
exists for it, and training falls back to mini-batch
gradient descent within a chunk, evaluated at a stale state, which is only accurate for small
chunks and, at such chunks, leaves accelerators mostly idle~\citep{zhang2026testtime, li2026tnt}. Recent work reduces the cost with very large
chunks~\citep{zhang2026testtime} or hierarchical memories~\citep{li2026tnt} but keeps the
stale mini-batch update. \tabref{tab:landscape} summarizes the landscape: existing memories are
WY-parallelizable as a whole or nonlinear; the closest precedents---the two-pass slot
memories ABC and GSA, and the layerwise delta-rule Delta MLP (\appref{sec:related})---write
their memories independently, none from the memory's own output error.

\begin{figure}[t]
  \centering
  \begin{tikzpicture}[
  font=\footnotesize,
  >={Stealth[length=4pt,width=3.5pt]},
  every node/.style={inner sep=1.5pt},
  mem/.style={draw=cSlotBorder, fill=cSlot, thick, rounded corners=2pt, minimum width=0.70cm, minimum height=0.70cm, inner sep=1pt},
  memwide/.style={mem, minimum width=0.88cm},
  memgray/.style={mem, fill=white, draw=cGray},
  act/.style={draw=cActBorder, fill=cAct, thick, circle, minimum size=0.46cm, inner sep=0pt},
  tgt/.style={draw=cTargetBorder, fill=cTarget, thick, rounded corners=2pt, minimum height=0.44cm, inner sep=2.5pt},
  flow/.style={->, thick, draw=cGray},
  back/.style={->, thick, draw=cTargetBorder, dashed},
  lbl/.style={text=cGray, font=\scriptsize},
  wy/.style={font=\scriptsize\itshape, text=cSlotBorder},
  title/.style={font=\footnotesize\bfseries, anchor=west},
  note/.style={lbl, anchor=north west, align=left},
]
\def\rowY{0.9}
\def\titleY{2.85}
\def\noteY{0.05}
\begin{scope}[shift={(0,0)}]
  \node[title] at (0,\titleY) {(a) Linear memory};
  \node[memwide] (S) at (1.75,\rowY) {$\mS_t$};
  \draw[flow] (0.55,\rowY+0.22) node[lbl,left] {$\vk_t$} -- ($(S.west)+(0,0.22)$);
  \draw[flow] (0.55,\rowY-0.22) node[lbl,left] {$\vv_t$} -- ($(S.west)+(0,-0.22)$);
  \draw[flow] (S.east) -- ++(0.6,0) node[lbl,right] {$\vo_t$};
  \node[wy, anchor=north] at (S.south) {WY kernel};
  \node[lbl, anchor=south, align=center] at ($(S.north)+(0,0.05)$) {one Householder\\step per token};
  \node[note, text width=3.6cm] at (0,\noteY)
    {\yes~exact chunkwise parallelism\\ \no~stores a linear map only};
\end{scope}
\begin{scope}[shift={(3.5,0)}]
  \node[title] at (0,\titleY) {(b) Deep memory (TTT-MLP/Titans)};
  \node[memgray] (W1) at (1.1,\rowY) {$\mW^{(1)}$};
  \node[act] (sig) at (2.05,\rowY) {$\sigma$};
  \node[memgray] (W2) at (3.0,\rowY) {$\mW^{(2)}$};
  \draw[flow] (0.35,\rowY) node[lbl,left] {$\vk_t$} -- (W1.west);
  \draw[flow] (W1.east) -- (sig.west);
  \draw[flow] (sig.east) -- (W2.west);
  \draw[flow] (W2.east) -- ++(0.55,0) node[lbl,right] {$\vhat_t$};
  \draw[back] ($(W2.north)+(0,0.05)$) |- ($(W1.north)+(0,0.5)$) -- ($(W1.north)+(0,0.05)$);
  \node[lbl, text=cTargetBorder, anchor=south] at ($(sig.north)+(0,0.55)$) {gradient through $\sigma$: no WY form};
  \node[wy, anchor=north, text=cGray] at (sig.south |- W1.south) {stale mini-batch update};
  \node[note, text width=3.9cm] at (0,\noteY)
    {\yes~nonlinear, wide memory\\ \no~small chunks, low utilization};
\end{scope}
\begin{scope}[shift={(8.8,0)}]
  \node[title] at (0,\titleY) {(c) \model};
  \node[memwide] (S1) at (1.2,\rowY) {$\Sone_t$};
  \node[act] (sig) at (2.35,\rowY) {$\sigma$};
  \node[memwide] (S2) at (3.5,\rowY) {$\Stwo_t$};
  \draw[flow] (0.35,\rowY) node[lbl,left] {$\vk_t$} -- (S1.west);
  \draw[flow] (S1.east) -- node[lbl, below, yshift=-1pt] {$\vz_t$} (sig.west);
  \draw[flow] (sig.east) -- node[lbl, below, yshift=-1pt] {$\vh_t$} (S2.west);
  \draw[flow] (S2.east) -- ++(0.55,0) node[lbl,right] {$\vhat_t$};
  \node[wy, anchor=north] at (S1.south) {WY kernel\,/\,apply-then-update};
  \node[wy, anchor=north] at (S2.south) {WY kernel};
  \node[tgt, draw=cGray, fill=white] (vv) at (3.5,\rowY+1.4) {value $\vv_t$};
  \draw[flow] (vv.south) -- (S2.north);
  \node[tgt] (tg) at (1.2,\rowY+1.4) {target $\vtil_t$};
  \draw[flow] (tg.south) -- (S1.north);
  \draw[back] (vv.west) -- node[lbl, text=cTargetBorder, below, align=center, yshift=-1pt] {error $\ve_t$\\projected back} (tg.east);
  \node[note, text width=5.0cm] at (0,\noteY)
    {\yes~nonlinear, wide memory ($\dhid$ is free)\\ \yes~each slot is a plain linear memory \mbox{(WY-executable)}};
\end{scope}
\end{tikzpicture}
  \caption{\textbf{\model at a glance.} (a)~A linear memory applies one Householder step per
  token, so a chunk of updates has an exact WY form. (b)~A deep memory is more expressive,
  but its gradient update through the nonlinearity has no such form. (c)~\model cascades two
  linear memories around a pointwise activation $\sigma$. The output error $\ve_t$ is
  projected back through the second memory into a target $\vtil_t$ for the first, so both
  layers are plain linear memories that the WY kernel can advance (by default the first uses
  the cheaper apply-then-update step of \algref{alg:toaster}); a chunk's targets and hidden
  keys are computed once, in parallel, from the chunk-initial states.}
  \label{fig:overview}
\end{figure}
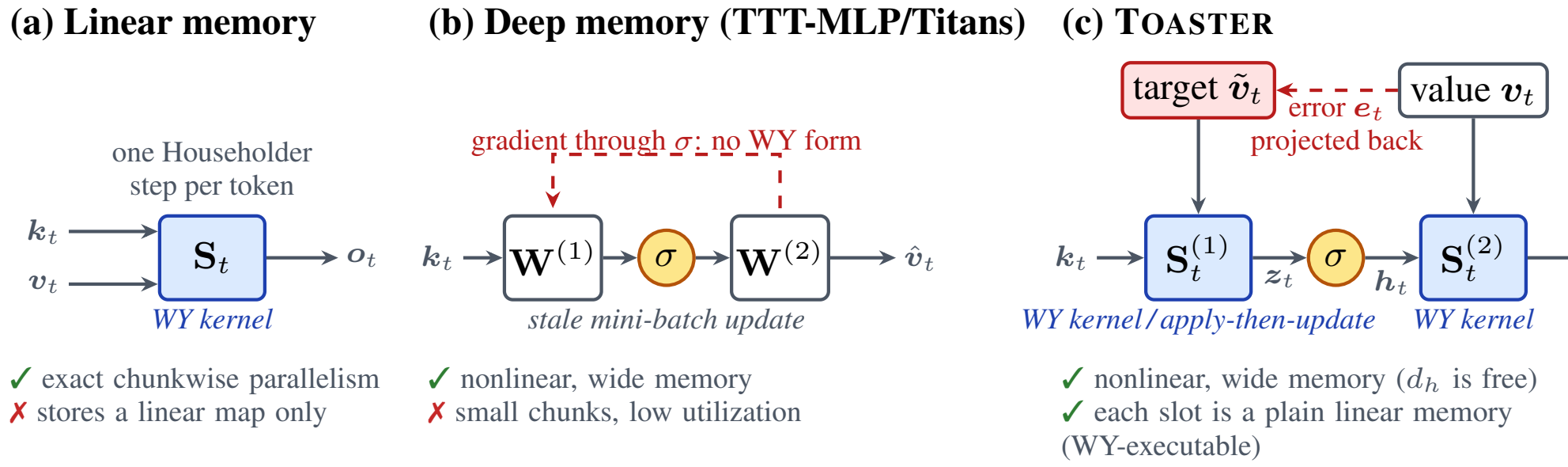
\suppressfloats[t]   %

\begin{table}[t]
  \centering
  \footnotesize
  \setlength{\tabcolsep}{4pt}
  \renewcommand{\arraystretch}{1.05}
  \begin{tabular}{@{}>{\raggedright\arraybackslash}p{1.38in}cc@{}}
    \toprule
    Memory & \shortstack{Exact chunkwise\\(WY) form} & \shortstack{Nonlinear\\state} \\
    \midrule
    \emph{Linear (delta rule):} DeltaNet, Gated DeltaNet, KDA, Gated DeltaNet-2, Comba,
    DeltaProduct & \yes & \no \\
    \emph{Deep (test-time training):} TTT-MLP, Titans, Atlas, LaCT, TNT & \no & \yes \\
    \midrule
    \textbf{\model} (this work) & \yes\ per layer$^\dagger$ & \yes \\
    \bottomrule
  \end{tabular}
  \caption{\textbf{Comparison between linear memory, deep memory, and \model.} $^\dagger$Given the chunk's cross-layer signals; by default the output memory uses the kernel and the hidden memory a summed chunk step (\secref{sec:practical}).}
  \label{tab:landscape}
\end{table}

\paragraph{This work.}
The two properties above turn out not to be in conflict, provided the cross-layer signals are approximated once per chunk, for a reason that lies in the delta rule
itself. A delta-rule write with key $\vk_t$ and value $\vv_t$ changes the state by $-\beta_t
\vk_t (\mS_{t-1}^\T \vk_t - \vv_t)^\T$: a rank-one update whose left factor is the key and
whose right factor is the residual between what is stored under the key and the value that replaces it. The gradient of any loss
with respect to a linear layer with input $\vk_t$ is also rank-one, $\vk_t \bm{g}_t^\T$, so, once the error $\bm{g}_t$ is given,
such a layer can be trained by an unmodified delta-rule kernel by handing it the value
$\vz_t - \bm{g}_t$, where $\vz_t$ is its current output and $\bm{g}_t$ the error arriving
there. \model applies this to a two-layer memory network $\vk \mapsto
\Stwo{}^{\T}\sigma(\Sone{}^{\T}\vk)$, i.e., two linear memories with a pointwise
activation between them (\figref{fig:overview}). The \emph{output} memory $\Stwo$ is
already an ordinary linear memory, with key $\vh_t = \sigma(\Sone{}^\T \vk_t)$ and value
$\vv_t$; the \emph{hidden} memory $\Sone$ needs a value, and projecting the output error
back through $\Stwo$ supplies one---a gradient-based target of the kind that target propagation uses for its top hidden layer in place of a learned inverse~\citep{lee2015difference}.
Once the hidden keys and targets of a chunk are
computed, each layer is a standard linear memory that an unmodified WY kernel can advance.
While the overall memory mapping is nonlinear, each state evolves using linear
updates.\footnote{Hence the name: two slots (linear memories) with heat (a nonlinearity) between
them; in principle nothing prevents more slots.} The precedents above write each memory on
its own; \model couples the two writes through the memory's own output error while each
layer keeps its kernel (\appref{sec:related}).

\paragraph{Contributions.}
\emph{(i)}~We introduce \model, a recurrent memory that cascades linear memories through
pointwise activations and trains the hidden layers by target propagation
(\secref{sec:method}). With the exact target and no forgetting, its per-token dynamics coincide with gradient
descent on a two-layer memory network; given the cross-layer signals of a chunk, each layer
can be advanced exactly by an existing WY kernel, so a chunk costs at most two kernel calls
plus a few dense matrix multiplications. \emph{(ii)}~We describe three design choices (\secref{sec:practical}). Two of them stabilized training across our scaling ladder: a bias through homogeneous coordinates, and a
bounded, rescaled back-projected error that also gives the hidden layer its smaller
step. We report the failure each one addressed. The third choice was made for throughput: a cheaper
apply-then-update step for the hidden memory in place of its kernel call. We ablate all
three at 145M (\tabref{tab:ablation}). \emph{(iii)}~Across a scaling ladder of five model sizes (0.2B--1.6B non-embedding
parameters) and two corpora, \model reaches lower language-modeling loss than KDA, the
linear memory it is built on, in all ten settings. In-domain, it matches or slightly improves on a
softmax-attention Transformer. At 1.2B, with all layers replaced, it is below Titans. In a 3:1 hybrid with global attention, it is
the best model on both corpora (0.18 and 0.44 perplexity
below attention on C4 and PG19). On a
13-task three-shot downstream suite, it has the highest average of the four models ($51.84\%$ against $50.52\%$ for KDA and $50.04\%$ for
Titans), and its hybrid has the highest average of the four hybrids ($52.11\%$). In length
extrapolation, its loss stays below KDA's up to 20K tokens, five times the training length
(\secref{sec:experiments}). \emph{(iv)}~At a hidden width matched to Titans' memory ($\dhid = \dk$), \model's training step is within $1.33\times$ of KDA's and $2.8$--$5.5\times$ faster than Titans at 32K (\figref{fig:throughput}).
Related work is discussed in \appref{sec:related}.

\section{Background: two kinds of recurrent memory}
\label{sec:background}

Given a token sequence with projections $\vq_t, \vk_t \in \R^{\dk}$ and $\vv_t \in
\R^{\dv}$, a recurrent memory maintains a state that is \emph{written} with $(\vk_t, \vv_t)$
and \emph{read} with $\vq_t$; a unifying view~\citep{sun2024learning, behrouz2026titans,
wang2025testtime} treats writing as one step of online optimization of a reconstruction loss
inside the forward pass. The two families below differ only in what the state is.

\paragraph{Linear memory and the WY representation.}
A linear memory stores a matrix $\mS_t \in \R^{\dk \times \dv}$, reads with $\vo_t =
\mS_t^\T \vq_t$, and writes by a gradient step on $\Loss_t(\mS) = \tfrac12 \lVert \mS^\T
\vk_t - \vv_t \rVert^2$ with step size $\beta_t \in (0,1]$, optionally preceded by a
forget gate $\alpha_t$:
\begin{equation}\label{eq:delta}
  \mS_t \;=\; \alpha_t \bigl(\I - \beta_t \vk_t \vk_t^\T\bigr)\, \mS_{t-1} \;+\; \beta_t\, \vk_t \vv_t^\T .
\end{equation}
With $\alpha_t \equiv 1$ this is DeltaNet~\citep{schlag2021linear};
a scalar data-dependent $\alpha_t$ gives Gated DeltaNet~\citep{yang2025gated}, and a
channel-wise gate $\Diag(\valpha_t)$, applied before the Householder factor, gives
KDA~\citep{team2025kimi} (\appref{app:wy}). What matters for training is that
\eqnref{eq:delta} is \emph{affine in the state}: the factor $\I - \beta_t \vk_t \vk_t^\T$
is a generalized Householder matrix, and over a chunk of $C$ tokens the product of the $C$
factors has the compact WY form $\I - \sum_{r=1}^{C} \vk_r \vw_r^\T$, with auxiliary
vectors $\vw_r$ obtained from a small triangular solve~\citep{bischof1987wy,
joffrain2006accumulating}. As a result, the state after the chunk and all $C$ outputs can be
computed \emph{exactly} from the state before the chunk with a few dense matrix
multiplications~\citep{yang2024parallelizing}; the recurrence is only sequential across
chunks. We write this chunkwise kernel as
\begin{equation}\label{eq:kernel}
  \bigl(\ch{\mS}{j+1},\; \ch{\mO}{j}\bigr) \;=\; \kernel\bigl(\ch{\mS}{j};\; \ch{\mK}{j}, \ch{\mV}{j}, \ch{\mQ}{j}\bigr),
\end{equation}
where $\ch{\mK}{j}, \ch{\mQ}{j} \in \R^{C \times \dk}$ and $\ch{\mV}{j}, \ch{\mO}{j} \in \R^{C \times \dv}$
stack the keys, queries, values, and outputs of chunk $j$, $\ch{\mS}{j}$ is the state
entering the chunk, and the gates $\alpha_t$ and step sizes $\beta_t$ of \eqnref{eq:delta} enter the kernel as further per-token inputs. More precisely, $\kernel$ applies
the $C$ delta-rule steps of \eqnref{eq:delta}, one per token in order, starting from
$\ch{\mS}{j}$. It returns the state $\ch{\mS}{j+1}$ after the last token and the $C$ outputs
$\ch{\mO}{j}$; the output of token $r$, $\vo_r = (\mS^{r})^\T \vq_r$, is read from the state
$\mS^{r}$ after that token. The result is identical to running the recurrence token by token; only the
arithmetic is reorganized into the dense products of the WY form. Efficient implementations
of $\kernel$ exist for all the gated variants above~\citep{yang2024fla}; \appref{app:wy}
gives a self-contained derivation. The limitation of this linear memory is its capacity: whatever it has seen, a linear
memory represents a single linear map $\vk \mapsto \mS^\T \vk$ with $\dk \dv$ numbers.

\paragraph{Deep memory and why the WY form is lost.}
Test-time-training memories~\citep{sun2024learning, behrouz2026titans, behrouz2025atlas}
replace the matrix by a neural network $\Mem(\cdot\,; \mathcal{W}_t)$, typically a
two-layer MLP, updated by gradient descent on the same reconstruction loss and read by a
forward pass, $\vo_t = \Mem(\vq_t; \mathcal{W}_t)$. The stored map is no longer linear, and
its expressivity is set by the width of the network, which also sets the size of its state, rather than by $\dk \dv$. But the gradient
with respect to the first-layer weights passes through the activation, so the update is not
affine in the state and no WY form exists. The standard remedy evaluates all gradients of a
chunk at the \emph{frozen} chunk-initial state~\citep{sun2024learning, behrouz2026titans},
\begin{equation}\label{eq:stale}
  \mathcal{W}_{[j+1]} \;=\; \alpha_{[j]}\, \mathcal{W}_{[j]} \;-\; \sum_{r=1}^{C}
  \beta_{[j],r}\, \nabla_{\mathcal{W}} \Loss_{[j],r}\bigl(\mathcal{W}_{[j]}\bigr),
\end{equation}
where $\mathcal{W}_{[j]}$ are the weights entering chunk $j$, $\Loss_{[j],r}(\mathcal{W}) =
\tfrac12 \lVert \Mem(\vk_r; \mathcal{W}) - \vv_r \rVert^2$ is the reconstruction loss of
token $r$ of the chunk, $\beta_{[j],r}$ is the step size of that token, and $\alpha_{[j]}$
is a chunk-level decay. For Titans this is a schematic, since Titans adds momentum and applies its decay per token~\citep{behrouz2026titans}. Every gradient in the sum is taken at the same state
$\mathcal{W}_{[j]}$, not at the state that the earlier tokens of the chunk would have
produced. This is
parallel within the chunk but is a mini-batch approximation whose error grows with $C$.
Implementations therefore use small chunks ($C$ between 16 and 64), pay a nonlinear forward
and backward pass per gradient, and run at a small fraction of peak
throughput~\citep{zhang2026testtime, li2026tnt}. Equation~\eqref{eq:stale} is also cruder
than \eqnref{eq:kernel} in kind: its gradients are frozen, so within a chunk no write sees the
writes before it (a per-token decay, as in Titans~\citep{behrouz2026titans}, can attenuate earlier writes but not correct them); our default hidden-layer update also takes this form. The next section shows that a deep memory can be
rearranged so that \emph{every layer} is a linear memory in its own right, recovering the
exact WY form layer by layer at the price of approximating the signals that pass between layers
once per chunk.

\section{\model}
\label{sec:method}

\model is a deep memory each of whose layers (excluding activations) is a linear memory. We define the cascade
(\secref{sec:cascade}), show how target propagation turns the hidden layer into a
linear-memory problem of its own (\secref{sec:tp}), give the chunkwise algorithm
(\secref{sec:chunk}), and describe the design choices that stabilized its training and make it
fast at scale (\secref{sec:practical}). We present two layers; \appref{app:general} treats $N$.

\subsection{A cascade of linear memories interleaved with nonlinear activations}
\label{sec:cascade}

\model keeps two matrix states, $\Sone_t \in \R^{\dk \times \dhid}$ and $\Stwo_t \in
\R^{\dhid \times \dv}$, each governed by a gated delta-rule recurrence of the form
\eqnref{eq:delta}. The hidden width $\dhid$ is a free hyperparameter: it sets the capacity of
the memory independently of the projection widths $\dk$ and $\dv$. A key $\vk_t$ is passed
through the cascade,
\begin{equation}\label{eq:forward}
  \vz_t = \Sone_{t-1}{}^{\T} \vk_t, \qquad
  \vh_t = \sigma(\vz_t), \qquad
  \vhat_t = \Stwo_{t-1}{}^{\T} \vh_t,
\end{equation}
where $\sigma$ is a pointwise activation (\figref{fig:overview}(c) draws this cascade). The composite map $\vk \mapsto \Stwo{}^\T
\sigma(\Sone{}^\T \vk)$ is a two-layer MLP with weights $\{\Sone_t, \Stwo_t\}$, the kind of
deep memory used by TTT-MLP and Titans. Reading uses the same cascade with the
query and the updated states, $\vo_t = \Stwo_t{}^\T \sigma(\Sone_t{}^\T \vq_t)$. The only
question is how to \emph{write}: how to update the two states so that the cascade as a
whole learns to map $\vk_t$ to $\vv_t$.

\subsection{Target propagation makes every layer a linear memory}
\label{sec:tp}

Both states $\Sone$ and $\Stwo$ should reduce the reconstruction loss $\Loss_t = \tfrac12 \lVert \vhat_t - \vv_t
\rVert^2$. For the \emph{output} layer this is already a linear-memory problem: $\Stwo$
maps $\vh_t$ to $\vhat_t$, and the delta rule of \eqnref{eq:delta} with key $\vh_t$ and
value $\vv_t$ is exactly a gradient step on $\Loss_t$ with respect to $\Stwo$.

The \emph{hidden} layer has no value of its own. Backpropagating $\Loss_t$ through
$\sigma$ and $\Stwo$ gives a gradient for $\Sone$, but as a function of the two states that
update is nonlinear, so the WY machinery does not apply to it directly. Target
propagation~\citep{bengio2014target, lee2015difference} asks a different question:
\emph{what value should $\vz_t$ have taken} for the loss to be smaller? We introduce a
target $\vtil_t \in \R^{\dhid}$ and give the hidden layer the local loss $\Loss^{(1)}_t =
\tfrac12 \lVert \vz_t - \vtil_t \rVert^2$, choosing $\vtil_t$ so that the local gradient,
taken with $\vtil_t$ held fixed, equals the true one, $\nabla_{\vz_t} \Loss^{(1)}_t = \nabla_{\vz_t} \Loss_t$. The local
gradient is $\vz_t - \vtil_t$. The true one follows from the chain rule through $\vhat_t =
\Stwo_{t-1}{}^{\T}\vh_t$ and $\vh_t = \sigma(\vz_t)$: $\nabla_{\vz_t} \Loss_t =
(\Stwo_{t-1}\ve_t) \had \sigma'(\vz_t)$, with $\ve_t = \vhat_t - \vv_t$ the output error
(\appref{app:target}). Equating the two and solving for $\vtil_t$ gives
\begin{equation}\label{eq:target}
  \boxed{\;
  \vtil_t \;=\; \vz_t \;-\; \bigl(\Stwo_{t-1}\, \ve_t\bigr) \had \sigma'(\vz_t),
  \qquad \ve_t = \vhat_t - \vv_t .
  \;}
\end{equation}
The target corrects the current pre-activation by the output error projected back through
the output memory and gated by the activation slope. With $\vtil_t$ in hand, the hidden
layer is updated by the delta rule with key $\vk_t$ and value $\vtil_t$---the same rule as
the output layer, and the same rule as any linear memory. No step size appears in
\eqnref{eq:target} because the target is a value, not an update: the step size is the
$\beta_t$ of that delta rule, \eqnref{eq:delta}, which sets how far the state moves toward
$\vtil_t$.

\begin{remark}[Per-token equivalence]\label{prop:equiv}
Let both layers use scalar step sizes $\beta^{(1)}_t, \beta^{(2)}_t$, no forgetting, and the
exact target \eqref{eq:target}, with $\vh_t$ and $\vtil_t$ computed from the previous
states. Updating $\Sone$ by \eqnref{eq:delta} with $(\vk_t, \vtil_t)$ and $\Stwo$ by
\eqnref{eq:delta} with $(\vh_t, \vv_t)$ is identical to one step of gradient descent on
$\Loss_t(\Sone, \Stwo)$ evaluated at $(\Sone_{t-1}, \Stwo_{t-1})$.
\end{remark}

The proof is two lines (\appref{app:target}): since $\nabla_{\Sone} \Loss_t = \vk_t (\vz_t
- \vtil_t)^\T$, the delta-rule step toward $(\vk_t, \vtil_t)$ \emph{is} the gradient step,
so token by token \model with the exact target is a two-layer memory of the TTT-MLP/Titans kind trained by
gradient descent. Notice that the joint transition of the two states is still nonlinear.
However, $\vh_t$ and $\vtil_t$ become \emph{keys and values that the kernel
consumes}, so that, conditioned on them, each state's transition is a Householder product.
This is the property the chunkwise algorithm exploits.

\subsection{Chunkwise-parallel training}
\label{sec:chunk}

\algref{alg:toaster} processes one chunk of $C$ tokens. Its inputs are the two states
entering the chunk, $\ch{\Sone}{j}$ and $\ch{\Stwo}{j}$, and the chunk's keys, queries, and
values, stacked as the rows of $\ch{\mK}{j}$, $\ch{\mQ}{j}$, and $\ch{\mV}{j}$. It has three
phases.

\emph{Phase~I} computes the cross-layer signals of every token at the frozen chunk-initial
states. It runs the forward pass of \eqnref{eq:forward} for all rows at once: the hidden
pre-activations $\ch{\mZ}{j} = \ch{\mK}{j}\ch{\Sone}{j}$, the hidden keys $\ch{\mH}{j} =
\sigma(\ch{\mZ}{j})$, and the output errors $\ch{\mE}{j} = \ch{\mH}{j}\ch{\Stwo}{j} -
\ch{\mV}{j}$. It then forms the hidden errors $\mathbf{G}_{[j]}$, the correction term of
\eqnref{eq:target_practical}: the output errors are projected back through
$\ch{\Stwo}{j}$, passed through the bounded normalization $\Norm$ of
\secref{sec:practical}, and gated by the activation slope. The
hidden targets are $\ch{\Vtil}{j} = \ch{\mZ}{j} - \mathbf{G}_{[j]}$. This phase is three
dense matrix products and a few pointwise operations.

\emph{Phase~II} advances the hidden memory with keys $\ch{\mK}{j}$ and values
$\ch{\Vtil}{j}$, and returns the hidden pre-activations $\ch{\mZ^{q}}{j}$ of the queries.
There are two ways to do this. Line~\ref{ln:wy} calls the chunkwise delta-rule kernel
\kernel{} of \eqnref{eq:kernel}. Line~\ref{ln:atu}, the default, uses apply-then-update:
every query reads the chunk-initial state, and the $C$ delta-rule writes are summed into one
update at the end of the chunk (\secref{sec:practical}).

\emph{Phase~III} advances the output memory with one call of \kernel{}, with the hidden
keys $\ch{\mH}{j}$ as keys, $\ch{\mV}{j}$ as values, and $\sigma(\ch{\mZ^{q}}{j})$ as
queries. It returns the chunk's outputs $\ch{\mO}{j}$. The two updated states are then
handed to the next chunk.

Given its keys
and values, a call of \kernel{} is exact for any chunk size. The approximations are that the
cross-layer signals are evaluated once, at the chunk-initial states, and, by default, that the
hidden layer sums its $C$ writes as in \eqnref{eq:stale} rather than applying them in
sequence. The first approximation is the price of the construction. For $C > 1$, the chunked algorithm takes one Jacobi (Picard) sweep across the layers and only approximates the gradient-descent memory of Remark~\ref{prop:equiv}, so the chunk size $C$ is part of the model (\tabref{tab:chunk_size}). Decoding with the training value of $C$ runs the same chunked recurrence and can be executed token by token with the same results up to rounding.

With no forgetting and the exact target, both update approximations vanish at $C = 1$, where the updates of
\algref{alg:toaster} reduce to Remark~\ref{prop:equiv}; both are first order in the
within-chunk change of the states (\appref{app:target}, which also gives the chunk update in
closed form).

We use $C = 32$, as for the linear baseline. With $\dhid = 4\dk$, a chunk costs
about seven times the arithmetic of the underlying kernel (nine times with \kernel{} on
both layers), yet the memory kernels remain a small part of a block: $8$--$9\%$ of its
multiply--accumulate operations at 1.2B, against $1\%$ for KDA (\appref{app:cost}). No new
kernel is required: our JAX implementation shares one chunkwise routine (\appref{app:wy}) with
the KDA baseline.

\begin{algorithm}[t]
\caption{\model, one chunk $j$. A subscript $[j]$ refers to chunk $j$: $\ch{\Sone}{j}$ and
$\ch{\Stwo}{j}$ are the states entering the chunk ($\ch{\Sone}{j+1}$ and $\ch{\Stwo}{j+1}$
the states leaving it), and each other subscripted matrix stacks its per-token quantity for
the $C$ tokens of the chunk, one row per token, as in \eqnref{eq:kernel}.
$\Norm$ is the bounded normalization of \secref{sec:practical}
(identity for the exact target) and $\vbeta^{(1)}$ the hidden-layer step sizes; gates are
omitted (passed to \kernel{} with the keys, \appref{app:wy}; averaged over the chunk on
line~\ref{ln:atu}).}
\label{alg:toaster}
\begin{algorithmic}[1]
\Require chunk-initial states $\ch{\Sone}{j}, \ch{\Stwo}{j}$; chunk inputs $\ch{\mK}{j}, \ch{\mQ}{j} \in \R^{C\times\dk}$, $\ch{\mV}{j} \in \R^{C\times\dv}$
\State \textbf{Phase I (cross-layer signals, all rows in parallel):}
\State \hspace{\algorithmicindent} $\ch{\mZ}{j} \gets \ch{\mK}{j} \ch{\Sone}{j}$;\quad
  $\ch{\mH}{j} \gets \sigma(\ch{\mZ}{j})$;\quad
  $\ch{\mE}{j} \gets \ch{\mH}{j} \ch{\Stwo}{j} - \ch{\mV}{j}$
\State \hspace{\algorithmicindent} $\mathbf{G}_{[j]} \gets \Norm\bigl(\ch{\mE}{j} \ch{\Stwo}{j}{}^{\T}\bigr) \had \sigma'(\ch{\mZ}{j})$;\quad
  $\ch{\Vtil}{j} \gets \ch{\mZ}{j} - \mathbf{G}_{[j]}$
  \Comment{hidden errors and targets}
\State \textbf{Phase II (hidden memory), one of:} \Comment{(b) is the default}
\State \hspace{\algorithmicindent} (a)~WY kernel:\quad
  $\bigl(\ch{\Sone}{j+1}, \ch{\mZ^{q}}{j}\bigr) \gets \kernel\bigl(\ch{\Sone}{j};\ \ch{\mK}{j}, \ch{\Vtil}{j}, \ch{\mQ}{j}\bigr)$ \label{ln:wy}
\State \hspace{\algorithmicindent} (b)~apply-then-update:\quad
  $\ch{\mZ^{q}}{j} \gets \ch{\mQ}{j}\,\ch{\Sone}{j}$;\quad
  $\ch{\Sone}{j+1} \gets \ch{\Sone}{j} - \ch{\mK}{j}{}^{\T}\Diag(\vbeta^{(1)})\,\mathbf{G}_{[j]}$ \label{ln:atu}
\State \textbf{Phase III (output memory):}
  $\bigl(\ch{\Stwo}{j+1}, \ch{\mO}{j}\bigr) \gets \kernel\bigl(\ch{\Stwo}{j};\ \ch{\mH}{j}, \ch{\mV}{j}, \sigma(\ch{\mZ^{q}}{j})\bigr)$
\State \Return $\ch{\Sone}{j+1}, \ch{\Stwo}{j+1}, \ch{\mO}{j}$
\end{algorithmic}
\end{algorithm}

\subsection{Design choices for stability and speed}
\label{sec:practical}

The target of \eqnref{eq:target} is exact but assumes ideal conditions: the hidden units
stay active rather than settling at zero, and the back-projected error keeps a stable scale.
Two modifications, one for each condition, stabilized the training of \model across our
scaling ladder. A third choice, made for
throughput rather than stability, selects the update rule of the hidden memory. We describe
each with the failure or cost it addressed in development. We ablate all three at 145M
(\tabref{tab:ablation}) and give the exact settings in \appref{app:hparams}.

\enlargethispage{\baselineskip}
\paragraph{A bias through homogeneous coordinates.}
Linear memories are usually written without a bias. In the cascade a constant input coordinate matters, because it shifts the pre-activations of the hidden layer: without a bias the pre-activations $\vz_t = \Sone{}^\T \vk_t$ pass through the
origin, and under ReLU-family activations many hidden units settle at zero. Because the state is rewritten at every token, the collapse
is self-reinforcing: in early runs without a bias, the effective rank of $\Sone$ collapsed
within a few hundred tokens, leaving a nearly linear cascade. We therefore absorb a bias into both states through homogeneous
coordinates~\citep{hartley2003multiple}, $\bar{\vk}_t = [\vk_t; 1]$ and $\bar{\vh}_t =
[\vh_t; 1]$, so that $\Sone_t \in \R^{(\dk+1)\times\dhid}$ and $\Stwo_t \in
\R^{(\dhid+1)\times\dv}$. Each layer remains an unmodified linear memory with one extra input
coordinate.

\paragraph{A bounded, rescaled back-projected error.}
In backpropagation the weights that transport the error change slowly; in \model the
transporting matrix $\Stwo_{t-1}$ is rewritten at every token and its norm can compound
over a long sequence, so the raw back-projection $\Stwo_{t-1}\ve_t$ produces targets whose
scale drifts, and the hidden memory diverges. We pass the back-projected error through a
bounded pointwise normalization before gating,
\begin{equation}\label{eq:target_practical}
  \vtil_t \;=\; \vz_t \;-\; \Norm\!\bigl(\Stwo_{t-1}\, \ve_t\bigr) \had \sigma'(\vz_t),
  \qquad \Norm(x) = \gamma\,\softsign(x) = \frac{\gamma\, x}{1 + \lvert x \rvert},
\end{equation}
with a fixed scale $\gamma = 0.01$. Each coordinate of the hidden correction is thus bounded
by $\gamma \max_z |\sigma'(z)|$ (about $1.13\gamma$ for GELU), so $\gamma$ also sets the effective step of the hidden layer relative to the
output layer, which is written with ground-truth values; the hidden layer needs a much
smaller step because its targets are back-projected through a $\dhid$-wide state. Since
$\softsign$ preserves the sign of each coordinate, the hidden update remains a descent
direction for $\Loss_t$ although it is no longer the gradient, provided it is taken per token, without forgetting, and with the residual evaluated at the current state (\appref{app:target}, which also states what this argument does not cover). In
preliminary runs, hard clipping and RMS or $\ell_2$ normalization of the whole vector were
less reliable across model sizes. Both memories otherwise inherit the input-dependent gates and step sizes of the
underlying KDA recurrence, with scales set separately from the baseline's
(\appref{app:hparams}). The initial states $\Sone_0$ and $\Stwo_0$ are learned, the
recurrent mixers use no positional encoding, and inference runs the same chunked recurrence
as training.

\paragraph{Apply-then-update for the hidden memory.}
The kernel on both layers is the form analyzed in \appref{app:target}, and it is the better
of the two hidden-memory update rules in \algref{alg:toaster}: at 145M it lowers C4
perplexity by $0.05$ relative to apply-then-update (15.64 against 15.69,
\tabref{tab:ablation}). It is also the slower one: in preliminary runs, training with the
kernel on both layers was $1.3$--$1.4\times$ slower per step than with the two dense products
of line~\ref{ln:atu}, which need no triangular solve. We use the same apply-then-update ordering as LaCT~\citep{zhang2026testtime} and TNT~\citep{li2026tnt}: every query of a chunk reads the chunk-initial state, and the chunk's writes take effect at its end. We trade that difference for
throughput: apply-then-update is the default for the hidden memory in every experiment unless
stated otherwise, with the kernel for the output memory; \figref{fig:throughput} uses these update rules at
a smaller hidden width, matched to Titans' memory (\appref{app:throughput}).

\paragraph{Summary: one chunk as run by default.}
With the three choices in place, a chunk runs as follows. Keys, queries, and hidden
activations carry an appended constant~1, so both states hold a bias. The bias coordinate
of the back-projected error is dropped (\appref{app:target}). Phase~I of
\algref{alg:toaster} computes, at the chunk-initial states, the hidden keys and targets of
all $C$ tokens, with the bounded back-projected error of \eqnref{eq:target_practical}.
Phase~II advances the hidden memory by apply-then-update (line~\ref{ln:atu}), with the gates
averaged over the chunk. Phase~III advances the output memory with one call of the WY kernel
\kernel{} and returns the chunk's outputs.

\section{Experiments}
\label{sec:experiments}

\begin{figure}[t]
  \centering
  \begin{minipage}[t]{0.48\textwidth}
    \centering
    \includegraphics{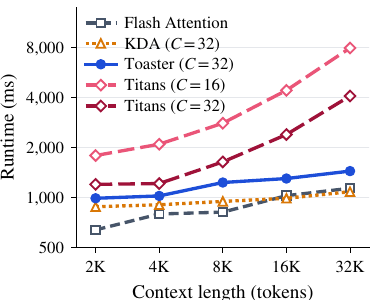}
    \caption{\textbf{Training-step time against context length} (145M, 512K tokens per step; hidden width $\dhid = \dk = 64$, matched to Titans' memory; \appref{app:throughput}).}
    \label{fig:throughput}
  \end{minipage}\hfill
  \begin{minipage}[t]{0.48\textwidth}
    \centering
    \includegraphics{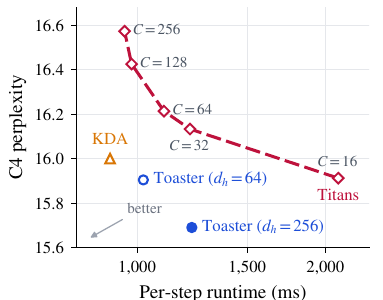}
    \caption{\textbf{Perplexity against per-step runtime} (145M, 4K context; $C$: chunk size).}
    \label{fig:ppl_speed}
  \end{minipage}
\end{figure}

\paragraph{Setup.}
\label{sec:setup}
We compare \model with softmax attention (\textbf{Attention}), the linear memory
\textbf{KDA}~\citep{team2025kimi}, and the deep test-time-training memory
\textbf{Titans}~\citep{behrouz2026titans}. We do not include Gated
DeltaNet~\citep{yang2025gated}, since KDA is a more expressive variant of it (channel-wise
rather than scalar forget gates) that consistently outperforms it~\citep{team2025kimi}. For a fair
comparison, all models use the decoder-only backbone of NanoDO~\citep{liu2024nanodo}, with
the attention block replaced by the corresponding sequence mixer, and are trained under
identical conditions on C4~\citep{raffel2020exploring} with a 4K-token context at five model sizes, labeled 145M to 1.2B (0.20B--1.58B non-embedding parameters, \tabref{tab:ladder_config}). Attention, KDA, and \model are trained along the whole ladder. Titans, whose training step is slower (\figref{fig:throughput}), is trained at 145M and 1.2B only. At 1.2B we also evaluate every mixer in the setting in which recurrent memories are usually deployed~\citep{team2025kimi, qwen2025qwen3next}: a \emph{hybrid} in which one global softmax-attention layer follows every three mixer layers, so that six of the 24 layers attend to the full context. This setting adds sliding-window attention (\textbf{SWA}, 128-token window) as a control mixer without a recurrent state. Model sizes and token budgets, optimizer settings, and
all architecture and mixer hyperparameters are given in \appref{app:hparams}; evaluation
settings are in \appref{app:eval}.

\begin{table}[t]
  \caption{\textbf{Validation perplexity across the scaling ladder} (lower is better; best per
  column in bold). \figref{fig:scaling} plots the C4 columns against training compute; the
  corresponding losses are in \appref{app:full_results}.}
  \label{tab:ladder_ppl}
  \centering
  \small
  \begin{tabular*}{\textwidth}{@{\extracolsep{\fill}} l ccccc ccccc @{}}
    \toprule
    & \multicolumn{5}{c}{C4 perplexity $\downarrow$ (in-domain)} & \multicolumn{5}{c}{PG19 perplexity $\downarrow$} \\
    \cmidrule(lr){2-6}\cmidrule(lr){7-11}
    Model & 145M & 300M & 550M & 850M & 1.2B & 145M & 300M & 550M & 850M & 1.2B \\
    \midrule
    \ladderpplwiderows
    \bottomrule
  \end{tabular*}
\end{table}

\begin{figure}[b]
  \centering
  \includegraphics{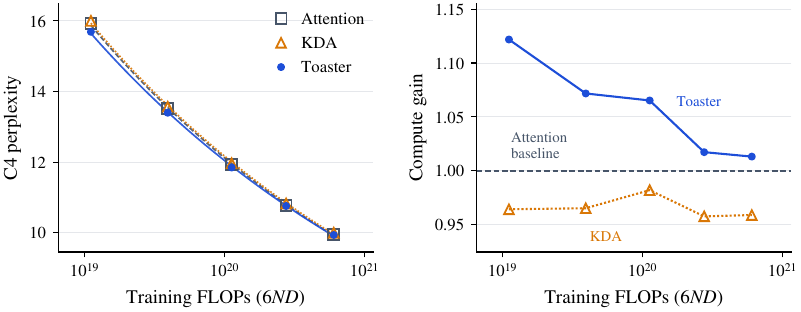}
  \caption{\textbf{Language modeling across the scaling ladder.} Left: the C4 perplexity of
  \tabref{tab:ladder_ppl} against training compute ($6ND$ FLOPs). Right: compute-efficiency
  gain relative to Attention at matched $6ND$ compute. Details and all numbers are in
  \appref{app:scaling_fit} and \appref{app:full_results}.}
  \label{fig:scaling}
\end{figure}

\paragraph{Language modeling across scales.}
\label{sec:results_lm}
\tabref{tab:ladder_ppl} and \figref{fig:scaling} summarize language-modeling quality along
the ladder. \model has the
lowest perplexity of the three architectures at every size on C4, up to and including 1.2B,
and it is below KDA at every size on both corpora. The differences at 1.2B are small in
absolute terms (0.02 perplexity over Attention and 0.06 over KDA on C4) but not trivial at
this scale, where the three architectures follow nearly parallel power laws in compute
(\figref{fig:scaling}, left) and a few hundredths of perplexity separate them. The right panel of
\figref{fig:scaling} makes this
concrete: on C4, Attention needs 12\% more compute than \model to reach the same loss at
145M and 1\% more at 1.2B, whereas KDA needs 2--4\% more compute than
Attention at every size. On PG19 the gain shrinks from 6\% at 145M to $-6\%$ at 1.2B. Our
hypothesis, which these runs do not test, is capacity: attention keeps the entire context available, whereas the memory
compresses it into a state whose size per head is fixed by $\dhid$ and whose total size grows only with the number of heads and layers (from 192 to 768 head--layer pairs along the ladder); as the models grow and the
rest of the network improves, this state may become the limiting factor
on long documents, so the memory capacity should be scaled together with the model, a
direction we leave to future work.
Compute is $6ND$, as in \citet{kaplan2020scaling}: it leaves out the sequence mixers' own
operations, whose share of a block falls as $1/D$ with width (for \model from 15\% at 145M to
8\% at 1.2B). Counting them raises the gains over Attention and offsets \model's gain over KDA
at these sizes (\appref{app:scaling_fit}).

\begin{table}[t]
  \caption{\textbf{Validation perplexity and three-shot downstream accuracy (\%) at 1.2B}, with
  all layers replaced by the sequence mixer (top) and in a 3:1 hybrid with global attention
  (bottom), in which one softmax-attention layer follows every three mixer layers; SWA is
  sliding-window attention with a 128-token window. Best and second best per column, within each
  group. Per-category averages and task details are in \appref{app:eval} and
  \appref{app:full_results}.}
  \label{tab:downstream}
  \centering
  \scriptsize
  \setlength{\tabcolsep}{2.1pt}
  \begin{tabular}{@{}l cc| cccccccccccccc@{}}
    \toprule
    \downstreammaingroups
    \cmidrule(lr){2-3}\cmidrule(lr){4-17}
    \downstreammainheader
    \midrule
    \downstreammainrows
    \bottomrule
  \end{tabular}
\end{table}

\paragraph{All layers replaced versus hybrid at 1.2B.}
\label{sec:results_hybrid}
\tabref{tab:downstream} compares the mixers at the largest size in two settings: with every layer
replaced by the mixer, and in the 3:1 hybrid with global attention. With all layers replaced,
\model has the lowest C4 perplexity of the four architectures. It is below Titans, the other
deep memory (hidden width 64, $C = 32$; \appref{app:mixers}), by 0.15 on C4 and 0.19 on PG19,
and KDA lies between the two on both corpora. On
PG19, Attention stays ahead of every recurrent model, as at the two largest sizes of the ladder. One attention layer
in four removes this deficit. Every hybrid is at or below Attention on both corpora. Even the
sliding-window hybrid, whose other 18 layers see only 128 tokens, matches Attention on PG19 and
is below it on C4, which shows how much of the attention model's quality six global layers
supply. Each recurrent memory improves on this control, and \model improves on it most. Its
hybrid has the lowest perplexity in the table, 0.18 below Attention on C4 and 0.44 below on
PG19, and 0.04 and 0.24 below the KDA hybrid. The mixers keep their order in the hybrid setting
(\model, then KDA, then Titans). The gain from the attention layers is larger on PG19
(0.38--0.54) than on C4 (0.16--0.19), and on PG19 it is largest for \model. This is consistent with the
capacity hypothesis above: a few global layers supply the long-range recall that a fixed-size
state lacks, and once they do, the advantage of the nonlinear memory shows in full.

\begin{figure}[b]
  \centering
  \includegraphics{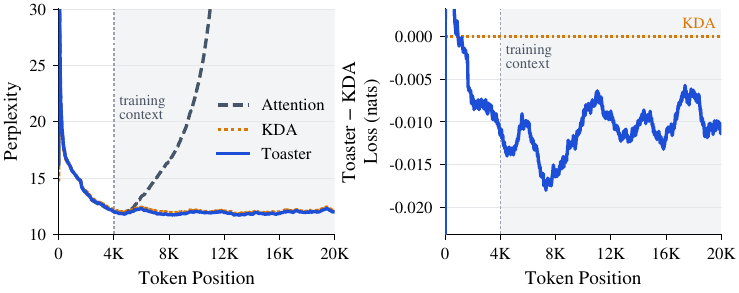}
  \caption{\textbf{Length extrapolation at 1.2B.} Left: perplexity by position. Right:
  \model's loss minus KDA's (negative favors \model). Curves are running averages over
  positions (\appref{app:extrapolation}).}
  \label{fig:extrapolation}
\end{figure}

\paragraph{Commonsense reasoning.}
\label{sec:results_downstream}
\tabref{tab:downstream} also reports three-shot accuracy at 1.2B on 13 commonsense-reasoning and language-understanding benchmarks (\appref{app:eval}).
With all layers replaced, \model has the \textbf{highest} average (51.8, against 50.7 for Attention, 50.5 for KDA, and 50.0 for Titans). It is at or above all three baselines on 9 of the 13 tasks, and no exception is larger than one point. Titans is 2--6 points below \model on ARC-Easy, HellaSwag, and PIQA. Six tasks are near chance for every model at this scale; on the other seven the averages are 57.7 for \model, 57.1 for KDA, 56.5 for Attention, and 56.1 for Titans (\appref{app:eval}). The hybrids keep this order, with the sliding-window hybrid in Attention's place: \model's hybrid has the highest average (52.1) and the best score on 8 of the 13 tasks, ahead of the sliding-window hybrid (51.6), KDA (50.9), and Titans (50.6). The six attention layers add less here than in perplexity, 0.3--0.5 points to each recurrent model's average, and the sliding-window hybrid, which has no recurrent state, comes second.

\paragraph{Length extrapolation on long sequences.}
All models are trained with a 4K context. \figref{fig:extrapolation} evaluates them on
20K-token sequences from the long-book question-answering split of
$\infty$Bench~\citep{zhang2024infinitebench}, five times the training length, reporting
perplexity by token position averaged over 100 sequences (\appref{app:extrapolation}).
Attention is evaluated directly at the longer length, without any length-extrapolation
technique such as position interpolation~\citep{chen2023extending},
YaRN~\citep{peng2024yarn}, or ALiBi~\citep{press2022train}; it degrades sharply beyond its
training context, whereas both recurrent models stay flat. \model's advantage over KDA is maintained over the entire range: beyond the first
thousand positions its loss stays below KDA's at every position, by $0.008$--$0.014$ nats
averaged over each 4K window, including the final window (16--20K, $0.009$ nats).

\newcommand{\niahgroups}{ & \multicolumn{4}{c}{S-NIAH-1 (passkey)} & \multicolumn{4}{c}{S-NIAH-2 (number)} & \multicolumn{4}{c}{S-NIAH-3 (UUID)} \\}
\newcommand{\niahmidrules}{\cmidrule(lr){2-5}\cmidrule(lr){6-9}\cmidrule(lr){10-13}}
\newcommand{\niahheader}{Model & 512 & 1K & 2K & 4K & 512 & 1K & 2K & 4K & 512 & 1K & 2K & 4K \\}
\newcommand{\niahrows}{%
Attention & 100.0 & 100.0 & 100.0 & 96.3 & 100.0 & \textbf{100.0} & \textbf{100.0} & \textbf{79.0} & \textbf{100.0} & \textbf{100.0} & \textbf{100.0} & \textbf{85.7} \\
KDA & 100.0 & 100.0 & 100.0 & \textbf{100.0} & 100.0 & 99.0 & 88.7 & 37.3 & 98.0 & 96.0 & 77.7 & 28.0 \\
\model & 100.0 & 100.0 & 100.0 & \textbf{100.0} & 100.0 & 99.3 & 90.3 & 62.7 & 98.0 & 97.0 & 73.7 & 27.3 \\}

\begin{table}[b]
  \caption{\textbf{Single needle-in-a-haystack accuracy (\%) at 1.2B}, averaged over 15
  needle positions, by context length (\appref{app:niah}).}
  \label{tab:niah}
  \centering
  \small
  \setlength{\tabcolsep}{4.2pt}
  \begin{tabular}{@{}l cccc cccc cccc@{}}
    \toprule
    \niahgroups
    \niahmidrules
    \niahheader
    \midrule
    \niahrows
    \bottomrule
  \end{tabular}
\end{table}

\paragraph{Needle in a haystack.}
We use the single-needle retrieval tasks of RULER~\citep{hsieh2024ruler}, adapted and
verified for our context lengths and tokenizer (\appref{app:niah}): a key--value pair (the
needle) is hidden in a long context (the haystack) and the model must return the value given
the key. S-NIAH-1 places a passkey in repeated synthetic text, and S-NIAH-2 and S-NIAH-3
place a number or a UUID in essay text, which tests whether the memory can retain the needle
while storing irrelevant context. \tabref{tab:niah} reports accuracy averaged over 15 needle
depths; the full pressure test, by depth and length, is in \appref{app:full_results}.
The recurrent models retrieve every passkey, and Attention retrieves at least 96\%. In the essay haystacks, the recurrent models lose needles that are
far from the query at 4K, whereas Attention, which keeps every token, retains 79--86\%.
\model retains more than KDA on S-NIAH-2 (63\% against 37\% at 4K) and matches it on the
UUID task (27\% against 28\% at 4K), whose random strings are the hardest content to keep in
a fixed-size state. In the 3:1 hybrids with global attention, \model retrieves 99.7\%, 89.3\%, and 77.3\% on S-NIAH-1, 2, and 3 at 4K, the highest of the four hybrids on each task (\tabref{tab:hybrid_niah}).

\paragraph{Throughput comparison.}
\model trains at close to the speed of the linear memory it is built from and, at long
context, several times faster than a deep memory of the same hidden width. \figref{fig:throughput}
reports the wall-clock time of one training step at a fixed budget of 512K tokens per step as
the context grows from 2K to 32K (\appref{app:throughput}). To compare training throughput at
matched fast-weight dimensions, \model and Titans are both
timed at hidden width $\dhid = \dk = 64$ (the width-matched configuration of \appref{app:throughput}), and all three recurrent models use 64-dimensional
heads. \model is $1.2$--$1.8\times$ faster than Titans at 2K and
$2.8$--$5.5\times$ faster at 32K, and the gap keeps widening with the context. Both memories evaluate a chunk's errors with one forward and backward pass at the chunk-initial state (Phase~I of \algref{alg:toaster} for \model). \model then advances its hidden memory with two dense products and its output memory with one call of the same chunkwise kernel as KDA (\secref{sec:chunk}). Titans' step time grows several-fold from 2K to 32K although every step processes the same number of tokens, whereas \model's stays within $1.33\times$ of KDA's at every context length and is
$1.27$--$1.55\times$ Flash Attention's, which KDA itself trails up to 8K. This overhead is
measured without any specialized kernel: \model reuses the KDA kernel unchanged for its output memory
and advances the hidden memory by the two dense products of apply-then-update
(\secref{sec:practical}); a fused kernel for the cascade is left to future work.

\paragraph{Perplexity versus training speed.}
\figref{fig:ppl_speed} compares perplexity and per-step runtime at 145M (timed as in
\figref{fig:throughput}). Titans evaluates all gradients of a chunk at the chunk-initial state
(\eqnref{eq:stale}), so its chunk size $C$ trades quality for speed: larger chunks train faster but
reach higher perplexity. At its best ($C = 16$), Titans is both slower and worse than \model, and
the chunk sizes that make it as fast as \model leave it worse than KDA. \model has the lowest
perplexity of the three architectures at a speed comparable to KDA's: nonlinear-memory quality at
close to linear-memory speed.

\begin{table}[t]
  \caption{\textbf{Ablations at 145M}: C4 perplexity and its difference $\Delta$ to the default \model
  (red: worse, blue: better); each column changes one component. Full results in \tabref{tab:ablation}
  (\appref{app:ablation}).}
  \label{tab:ablation_main}
  \centering
  \small
  \setlength{\tabcolsep}{3.5pt}
  \begin{tabular}{@{}l cc cc cc cc c@{}}
    \toprule
    & \multicolumn{2}{c}{Reference} & \multicolumn{2}{c}{State width} & \multicolumn{2}{c}{Activation $\sigma$}
    & \multicolumn{2}{c}{Hidden-layer learning} & Bias \\
    \cmidrule(lr){2-3}\cmidrule(lr){4-5}\cmidrule(lr){6-7}\cmidrule(lr){8-9}\cmidrule(l){10-10}
    & KDA & \model & $\dhid = \dk$ & $\dhid = 2\dk$ & identity & leaky ReLU & $\gamma = 0$ & WY kernel & w/o \\
    \midrule
    C4 ppl ($\downarrow$) & 16.00 & 15.69 & 15.90 & 15.76 & 15.80 & 15.71 & 15.71 & 15.64 & 15.86 \\
    $\Delta$ & \cellcolor{red!45}$+$0.31 & -- & \cellcolor{red!39}$+$0.22 & \cellcolor{red!13}$+$0.07
    & \cellcolor{red!20}$+$0.11 & \cellcolor{red!4}$+$0.02 & \cellcolor{red!3}$+$0.02
    & \cellcolor{blue!9}$-$0.05 & \cellcolor{red!31}$+$0.17 \\
    \bottomrule
  \end{tabular}
\end{table}

\paragraph{Ablation study.}
\label{sec:ablation}
\tabref{tab:ablation} varies one component of \model at a time at 145M (\appref{app:ablation});
apart from the slower WY-kernel update of the hidden memory, a wider hidden layer, and other chunk sizes (\appref{app:additional}), no change lowers perplexity by more
than 0.01. Three comparisons in \tabref{tab:ablation_main} (single runs) locate the 0.31 gain over KDA: a larger state, from which an affine cascade
with the same state (identity activation) already gains 0.20, although that row also differs from KDA in its bias, learned initial states, and gate scales; the nonlinearity, which adds a
further 0.11 (GELU is the best of six activations); and target propagation, which lets the
hidden memory learn in context (0.02 over $\gamma = 0$; advancing the hidden memory exactly with
the WY kernel lowers perplexity by a further 0.05). Most of the gain thus comes from a wide nonlinear memory that runs on linear-memory kernels
(full discussion in \appref{app:ablation}). Further ablations are in \appref{app:additional}: with the hidden memory $\Sone$ completely frozen, perplexity is 15.86, against 15.71 at $\gamma = 0$ and 15.69 for the default (\tabref{tab:frozen_hidden}); KDA with $8\times$ its default state reaches 15.79 (\tabref{tab:kda_state}); training chunk sizes from 8 to 256 give 15.56--15.74 (\tabref{tab:chunk_size}); and $\dhid = 512$ gives 15.62, whereas a softmax activation gives 18.19 (\tabref{tab:ablation}).

\section{Conclusion and limitations}
\label{sec:conclusion}

\model bridges two families of recurrent memory. By writing
a two-layer memory as two linear memories with an activation in between, and giving the
hidden memory its own targets through target propagation, it lets every layer keep a linear memory's Householder transition (by default only the output memory keeps it; the hidden memory takes the cheaper apply-then-update step) and confines the nonlinearity to the choice of
what is written: with the exact target and no forgetting it is, token by token, gradient descent on a
two-layer memory network, and chunk by chunk it is at most two calls to an unmodified linear-memory kernel, given cross-layer signals that are approximated once per chunk. On a
five-point scaling ladder it improves on the linear memory it is built from in every setting,
by a margin that narrows with scale, and is on par with softmax attention in-domain; in a 3:1 hybrid with global attention at 1.2B it improves on attention on both corpora. The recipe---keep every layer's transition
affine, move the nonlinearity into the targets---applies in principle to any linear
recurrence with a chunkwise kernel.

\paragraph{Limitations.}
We would like to note five limitations. First, the efficiency evidence is an operation count
(\appref{app:cost}) and one throughput profile (\figref{fig:throughput}), measured at a
single model size on one platform, with no specialized kernel for \model and no LaCT
baseline; with $\dhid = 4\dk$ \model has about eight times the state and seven to nine times
the kernel arithmetic of KDA, and the compute-efficiency gains of \figref{fig:scaling} are per
training token: by the operation counts of \appref{app:cost}, \model costs $6$--$13\%$ more
per token than KDA (up to $16\%$ with the kernel on both layers), which offsets its gain over KDA at equal compute (\tabref{tab:mixer_cost}). Second, all training runs are
single runs with one seed, the study stops at 1.6B non-embedding parameters, 63.7B tokens,
and a 4,096-token context, and the KDA baseline uses 64-dimensional heads; larger scales,
wider-head or state-matched linear baselines beyond the 145M ablation of \tabref{tab:kda_state}, and recall beyond 4K are untested. Third, the
downstream evidence is thin: on the 13-task three-shot suite \model has the best average by
about a point, but we ran no per-task significance tests and six tasks are at or near chance
for every model. Fourth, the design choices of \secref{sec:practical} are ablated only
at 145M (\tabref{tab:ablation}), KDA has not been retuned with \model's gate scales, and the
effect of the chunk size on the approximation of \secref{sec:chunk} is measured only at 145M (\tabref{tab:chunk_size}).
Finally, we evaluated depth two only; deeper cascades (\appref{app:general}) are untested.

\bibliography{references}
\bibliographystyle{iclr2027_conference}

\newpage
\appendix
\raggedbottom   %
\section{Statements}
\label{app:statements}

\subsection{Reproducibility statement}

All architectural details needed to implement \model are given in \secref{sec:method}
(forward pass, target, and chunkwise algorithm) and \appref{app:general} (deeper cascades),
with the target derivation and the per-token equivalence proof in \appref{app:target} and
the WY algorithm behind the kernel in \appref{app:wy}. \appref{app:hparams} lists the
complete scaling ladder (depth, width, heads, batch size, learning rate, schedule, and token
budget per size), the optimizer settings, the shared backbone, the mixer-specific settings
of KDA and \model, and the tokenizer; \appref{app:eval} describes the three-shot downstream evaluation
protocol; and \appref{app:full_results} reports every number behind the figures and
summary tables of the main text. Source code for the \model layer and the evaluation
configuration will be released upon publication.

\subsection{Ethics statement}

This work studies the architecture of language-model memory, trains models on the public
web corpus C4, and evaluates them on C4, PG19, $\infty$Bench, and public benchmark suites. It involves no human subjects, and it neither
creates nor releases new datasets. The risks and benefits of the resulting models are those
of language models trained on public web text in general; more efficient long-context
memory lowers the cost of such models and may broaden both beneficial and harmful uses in
equal measure. We are not aware of ethical concerns specific to this work.

\subsection{AI use statement}

We used generative AI tools to draft and edit the text, to summarize related literature, to
create figures, and to typeset the derivations in \appref{app:target} and \appref{app:wy}. We
also used them to write the code that generates tables and plots from experiment logs, to
edit code in the evaluation pipeline, to help interpret the experimental results, and for
feedback on the experimental design, on the mathematical statements of \appref{app:target}, and
on the proof of Proposition~\ref{prop:chunk}. We did not use them for research ideation, for
the design of the method, for the mathematical formulation of the target-propagation update,
for the implementation of the training and inference code, for synthetic data generation, or
to run experiments. We checked all AI-assisted text, derivations, numbers, and related-work
claims, and we take responsibility for the final content of this work.
\section{Related work}
\label{sec:related}

\paragraph{Linear recurrent memories.}
Linear attention~\citep{katharopoulos2020transformers} and its gated and state-space
descendants~\citep{sun2024retentive, yang2024gated, gu2024mamba, dao2024transformers,
peng2025rwkv7, lahoti2026mamba} maintain a matrix state updated by an outer-product write.
The delta rule replaces the pure write with an erase-and-write Householder
step~\citep{schlag2021linear}, and \citet{yang2024parallelizing} showed that the WY
representation~\citep{bischof1987wy, joffrain2006accumulating} yields an exact, matmul-rich
chunkwise algorithm; Gated DeltaNet~\citep{yang2025gated}, KDA~\citep{team2025kimi}, Gated
DeltaNet-2~\citep{hatamizadeh2026gdn2}, Comba~\citep{hu2025comba},
DeltaProduct~\citep{siems2025deltaproduct}, preconditioned
variants~\citep{tumma2026preconditioned}, and MesaNet~\citep{vonoswald2026mesanet} refine
the gate, rank, geometry, or optimality of this step while keeping a chunkwise form. These memories now run in production hybrids that
interleave three linear layers with one softmax-attention layer: Qwen3-Next with Gated
DeltaNet~\citep{qwen2025qwen3next}, and Kimi Linear and the 2.8T-parameter Kimi K3 with
KDA~\citep{team2025kimi, kimik3}. Notably, all of K3's recurrent layers are linear
delta-rule memories computed with WY/UT-transform kernels and the report contains no
nonlinear memory. \secref{sec:experiments} evaluates \model in this 3:1 layout (\tabref{tab:downstream}).

\paragraph{Two-pass slot memories and deep fast-weight memories.}
ABC~\citep{peng2022abc} bounds attention's memory to a fixed number of slots, storing two
matrices that are read through a softmax over the slot keys, and GSA~\citep{zhang2024gsa}
gates the slot updates so that, in the authors' words, it ``comprises a two-layer GLA linked
via softmax'' and is trained as two chunkwise passes. Raven~\citep{afzal2026raven} writes its slot memory through a learned sparse router and also reads it through a softmax over slots. These are the closest precedents for
placing a nonlinearity between two linear memories that each keep a chunkwise kernel; the
two memories are written independently, from the input alone (through gating in GSA and a learned sparse router in Raven), so that no error signal from the output reaches the first memory, and GSA notes that the
analogous two-pass DeltaNet was not explored. Delta MLP~\citep{irie2021going} writes every
layer of a deep fast-weight network with the delta rule, from keys and values produced by the
slow network, and the metalearned neural memory~\citep{munkhdalai2019metalearned} updates
each layer toward targets produced by a learned feedback network; in neither is a layer's
target derived from the output error of the memory itself. \model differs from all five in
that its two writes are coupled: the output memory is keyed by the hidden activations, and
the hidden memory is written toward a target derived from the output error, so that with the
exact target the pair is one gradient step on the composite memory
(Remark~\ref{prop:equiv}).

\paragraph{Nonlinear (test-time-training) memories.}
Fast-weight memories written at test time go back to \citet{ba2016using}, deep and recurrent
fast-weight programmers to \citet{irie2021going}, and deep memories with locally computed
targets to the metalearned neural memory of \citet{munkhdalai2019metalearned}.
TTT-MLP~\citep{sun2024learning} and Titans~\citep{behrouz2026titans} store an MLP updated by
gradient descent at test time; Miras~\citep{behrouz2025miras}, Atlas~\citep{behrouz2025atlas},
Nested Learning~\citep{behrouz2025nested}, memory caching~\citep{behrouz2026memorycaching},
and end-to-end or in-place variants~\citep{tandon2025e2e, feng2026inplace} broaden the
objectives, optimizers, and placement of such memories. Their training cost has been
attacked from two directions: LaCT~\citep{zhang2026testtime} uses very large chunks so that
each update is a large matmul, and TNT~\citep{li2026tnt} splits the memory into a global
module updated on large chunks and local modules updated in parallel. Both keep the stale
mini-batch update of \eqnref{eq:stale}. Concurrently, E$^2$-TTT~\citep{zhong2026e2ttt}
derives closed-form chunk transitions that reproduce the exact chunk-end weights and momentum
of an MLP memory, with every per-token gradient still evaluated at the chunk-start weights;
\model instead removes the obstacle to exact chunkwise computation within each layer, by
making every layer a linear memory, and, with the kernel on both layers, freezes only the
cross-layer signals at the chunk start (Proposition~\ref{prop:chunk}); its default advances
the hidden layer by the summed chunk-start step of \eqnref{eq:stale}, like these methods,
and the output layer exactly. Modular TTT~\citep{tang2026modular} ablates the
components of TTT and finds that deeper fast-weight networks tend to hurt at the scales
tested, which it attributes to excessively large activations; \model writes its hidden layer
with a bounded, separately scaled correction rather than a raw gradient
(\secref{sec:practical}), and the cascade improves on the one-layer memory at every size of
our ladder. Complementary to ours, \citet{wang2025testtime} frame linear and
test-time-training memories as test-time regression, and \citet{liu2026tttla} prove that
TTT with key--value binding is linear attention with a learned, state-dependent feature map;
\model's read, $\Stwo{}^{\T}\sigma(\Sone{}^{\T}\vq)$, is of this form, so the question is
whether writing the hidden memory in context adds anything beyond a learned nonlinear key
map, which the identity and $\gamma = 0$ rows of \tabref{tab:ablation} and the frozen-memory run of \tabref{tab:frozen_hidden} test.

\paragraph{Target propagation.}
Target propagation~\citep{bengio2014target, lee2015difference} trains deep networks by
giving each layer a local target instead of a backpropagated gradient; later work analyzed
it as approximate Gauss--Newton~\citep{meulemans2020theoretical}, applied it to recurrent
networks~\citep{roulet2023target}, and learned or fixed the feedback
pathways~\citep{ernoult2022towards, shibuya2023fixed}. Our purpose differs: we seek not a
backprop-free learning rule but a \emph{form} of the exact gradient step in which each
layer's update is a standard linear-memory write, because that is what the WY kernel can
execute. The observation behind \eqnref{eq:target}, that a gradient step on a linear layer
is a key--value write with the input as key and the error as value, is the dual form of
gradient-trained layers of \citet{irie2022dual}; and the target $\vz_t - \bm{g}_t$, a
pre-activation moved by one gradient step and then fitted with a local squared loss, is the
local-target construction of ProxProp~\citep{frerix2018proxprop} and
LocoProp~\citep{amid2022locoprop}, which \model applies inside a recurrent memory, where
the local fit is executed by a delta-rule kernel. Related in spirit, \citet{cerruti2026linear} forward a lower layer's delta-rule
write error or write value into the next layer of a network; \model routes the analogous
signal inside a single memory, between its two slots.
\section{Deeper cascades}
\label{app:general}

The construction of \secref{sec:method} extends to a cascade of $N$ linear memories
$\mS^{(n)}_t \in \R^{d_{n-1} \times d_n}$, $n = 1, \dots, N$, with $d_0 = \dk$ and $d_N =
\dv$. Writing $\vh^{(0)}_t = \vk_t$, the forward pass is
\begin{equation}
  \vz^{(n)}_t = \mS^{(n)}_{t-1}{}^{\T} \vh^{(n-1)}_t, \qquad
  \vh^{(n)}_t = \sigma(\vz^{(n)}_t) \;\; (n < N), \qquad
  \vhat_t = \vz^{(N)}_t ,
\end{equation}
so that the output layer has no activation. Let $\Loss_t = \tfrac12\lVert \vhat_t - \vv_t
\rVert^2$ and define, for every layer, the target $\vtil^{(n)}_t$ through
$\nabla_{\vz^{(n)}_t} \Loss_t = \vz^{(n)}_t - \vtil^{(n)}_t$. The chain rule gives the
backward recursion
\begin{equation}\label{eq:general_target}
  \vtil^{(N)}_t = \vv_t, \qquad
  \vtil^{(n)}_t = \vz^{(n)}_t - \Bigl(\mS^{(n+1)}_{t-1}\bigl(\vz^{(n+1)}_t - \vtil^{(n+1)}_t\bigr)\Bigr) \had \sigma'(\vz^{(n)}_t)
  \quad (n = N-1, \dots, 1),
\end{equation}
which for $N = 2$ is \eqnref{eq:target}. Layer $n$ is then written with key
$\vh^{(n-1)}_t$ and value $\vtil^{(n)}_t$, and, exactly as in Remark~\ref{prop:equiv},
the $N$ delta-rule steps coincide with one gradient step on $\Loss_t$ with respect to all
$N$ states. The chunkwise algorithm generalizes accordingly: Phase~I runs the forward
recursion and then \eqnref{eq:general_target} on the frozen chunk-initial states, producing
all keys and targets of the chunk; Phases~II--$(N{+}1)$ advance the layers in order
$n = 1, \dots, N$, each hidden layer by either rule of \algref{alg:toaster} and the output
layer by \kernel{}, each phase taking as queries the activated per-query outputs of the
previous one. The cost is at most $N$ sequentially dependent kernel calls and $O(N)$ dense products per chunk. In
practice the bounded normalization $\Norm$ of \eqnref{eq:target_practical} is applied to
each back-projected error in \eqnref{eq:general_target}, and the frozen-target approximation
of \secref{sec:chunk} occurs at every hidden layer. We have not trained cascades with $N > 2$.

\section{Target derivation and proof of Remark~\ref{prop:equiv}}
\label{app:target}

\paragraph{Derivation of \eqnref{eq:target}.}
With the forward pass \eqnref{eq:forward} and $\ve_t = \vhat_t - \vv_t$, the chain rule
through $\vhat_t = \Stwo_{t-1}{}^{\T}\vh_t$ and $\vh_t = \sigma(\vz_t)$ gives
\begin{equation}
  \nabla_{\vz_t} \Loss_t
  = \Bigl(\frac{\partial \vh_t}{\partial \vz_t}\Bigr)^{\!\T}
    \Bigl(\frac{\partial \vhat_t}{\partial \vh_t}\Bigr)^{\!\T} \nabla_{\vhat_t}\Loss_t
  = \Diag\bigl(\sigma'(\vz_t)\bigr)\, \Stwo_{t-1}\, \ve_t
  = \bigl(\Stwo_{t-1}\, \ve_t\bigr) \had \sigma'(\vz_t),
\end{equation}
where we used the Jacobians $\partial \vh_t / \partial \vz_t = \Diag(\sigma'(\vz_t))$ and
$\partial \vhat_t / \partial \vh_t = \Stwo_{t-1}{}^{\T}$. The local loss $\Loss^{(1)}_t = \tfrac12 \lVert \vz_t - \vtil_t \rVert^2$ has
gradient $\vz_t - \vtil_t$ when $\vtil_t$ is held fixed; equating the two gradients and solving for $\vtil_t$ yields
\eqnref{eq:target}.

\paragraph{Proof of Remark~\ref{prop:equiv}.}
The gradients of $\Loss_t$ with respect to the two states, evaluated at
$(\Sone_{t-1}, \Stwo_{t-1})$, are
\begin{equation}
  \nabla_{\Stwo} \Loss_t = \vh_t\, \ve_t^{\T}, \qquad
  \nabla_{\Sone} \Loss_t = \vk_t\, \bigl(\nabla_{\vz_t}\Loss_t\bigr)^{\T} = \vk_t\,(\vz_t - \vtil_t)^{\T},
\end{equation}
because $\vhat_t$ is linear in $\Stwo$ with input $\vh_t$, and $\vz_t$ is linear in $\Sone$
with input $\vk_t$. The delta rule \eqnref{eq:delta} without forgetting, applied to $\Stwo$
with key $\vh_t$ and value $\vv_t$, reads
\begin{equation}
  \Stwo_t = \Stwo_{t-1} - \beta^{(2)}_t \vh_t \bigl(\Stwo_{t-1}{}^{\T}\vh_t - \vv_t\bigr)^{\T}
          = \Stwo_{t-1} - \beta^{(2)}_t\, \vh_t\, \ve_t^{\T}
          = \Stwo_{t-1} - \beta^{(2)}_t\, \nabla_{\Stwo}\Loss_t ,
\end{equation}
and applied to $\Sone$ with key $\vk_t$ and value $\vtil_t$,
\begin{equation}
  \Sone_t = \Sone_{t-1} - \beta^{(1)}_t \vk_t \bigl(\Sone_{t-1}{}^{\T}\vk_t - \vtil_t\bigr)^{\T}
          = \Sone_{t-1} - \beta^{(1)}_t\, \vk_t\, (\vz_t - \vtil_t)^{\T}
          = \Sone_{t-1} - \beta^{(1)}_t\, \nabla_{\Sone}\Loss_t .
\end{equation}
Both updates are gradient steps on $\Loss_t$ evaluated at the previous states, which is
the claim. With a forget gate $\alpha_t$ the same computation shows that the output-layer step is a gradient step taken after weight decay; for the hidden layer, whose target is computed from the undecayed state, the gated delta rule adds $\beta^{(1)}_t(1-\alpha_t)\,\vk_t\vz_t^{\T}$ to that step, partly undoing the decay along $\vk_t$ (Titans, by contrast, evaluates the
gradient before its decay and adds momentum). \qed

\paragraph{What the chunkwise algorithm computes.}
In \algref{alg:toaster}, Phase~I is three dense matrix multiplications and a few pointwise
operations, fully parallel across the chunk. Phase~II advances the hidden memory, with keys
$\ch{\mK}{j}$ and values $\ch{\Vtil}{j}$, and returns the per-query pre-activations
$\ch{\mZ^{q}}{j}$, either by a call of \kernel{} (line~\ref{ln:wy}) or, by default, by
apply-then-update (line~\ref{ln:atu}): every query of the chunk reads the chunk-initial state,
and the $C$ delta-rule writes, whose residuals $\mathbf{G}_{[j]}$ Phase~I has already
evaluated at that state, are summed into one update at the end of the chunk. Phase~III calls
\kernel{} for the output memory, with keys $\ch{\mH}{j}$, values $\ch{\mV}{j}$, and queries
$\sigma(\ch{\mZ^{q}}{j})$, returning the chunk's output. With the homogeneous coordinates of
\secref{sec:practical}, keys, queries, and hidden activations carry an appended constant~1,
and the bias coordinate of the back-projection is dropped. Given the hidden keys
$\ch{\mH}{j}$ and targets $\ch{\Vtil}{j}$, a kernel call computes its layer's delta-rule
recurrence exactly, for any chunk size: within the chunk, later writes act on the state left by
earlier ones, and the layer's residual is evaluated at its running state rather than at the
chunk-initial one, which \eqnref{eq:stale} cannot do. The approximation is that the
cross-layer signals are evaluated once, at the chunk-initial states. We do not claim that
this approximation is uniformly smaller than that of \eqnref{eq:stale}; what a kernel call
removes is the additive update with frozen residuals within its layer. For the output layer the
approximation only stales the keys $\vh_r$; for the hidden layer its effect depends on the
update rule, as the next paragraph shows.

\paragraph{The chunkwise hidden update.}
In \algref{alg:toaster} with the kernel on the hidden layer (line~\ref{ln:wy}), the target of
token $r$ in chunk $j$ is $\vtil_r = \vz_r^{[j]} -
\bm{g}_r^{[j]}$, where $\vz_r^{[j]} = \ch{\Sone}{j}{}^{\T}\vk_r$ and $\bm{g}_r^{[j]}$ are the
pre-activation and back-projected error at the chunk-initial states. Writing $\mS^{(1),r}$
for the hidden state after token $r$ of the chunk, the kernel applies the delta rule with the
running state, so the executed update is
\begin{equation}
  \mS^{(1),r} - \mS^{(1),r-1}
  = -\beta_r \vk_r \bigl(\mS^{(1),r-1}{}^{\T}\vk_r - \vtil_r\bigr)^{\T}
  = -\beta_r \vk_r \bigl(\vz_r - \vz_r^{[j]} + \bm{g}_r^{[j]}\bigr)^{\T},
\end{equation}
with $\vz_r = \mS^{(1),r-1}{}^{\T}\vk_r$ the running pre-activation. The term $\vz_r - \vz_r^{[j]} = (\mS^{(1),r-1} - \ch{\Sone}{j})^{\T} \vk_r$ is the change in
the hidden layer's prediction on $\vk_r$ produced by the earlier tokens of the chunk; it is
small when those tokens' keys are nearly orthogonal to $\vk_r$ (with the homogeneous
coordinate the inner products are $\vk_i^{\T}\vk_r + 1$, so it does not vanish exactly),
vanishes at $C = 1$, and is first order in the within-chunk change of the state. With apply-then-update (line~\ref{ln:atu}) the residual
of every token is evaluated at $\ch{\Sone}{j}$, where the target was computed, so the
executed per-token update is exactly $-\beta_r \vk_r (\bm{g}_r^{[j]})^{\T}$ and the $C$
updates are summed, as in \eqnref{eq:stale}; the term $\vz_r - \vz_r^{[j]}$ does not arise,
and the queries of the chunk read $\ch{\Sone}{j}$.

\paragraph{The chunk update in closed form.}
Proposition~\ref{prop:chunk} states what \algref{alg:toaster} computes over a whole chunk
with the kernel on both layers, when the forget gates are off and the exact target
\eqnref{eq:target} is used, and compares
it with chunked gradient descent at the chunk-initial state, i.e., \eqnref{eq:stale} without
decay. For a matrix $\mathbf{X} \in \R^{C \times d}$ with rows $\vx_r$ and step sizes $\vbeta$, write
$\mT_{\mathbf{X}} = \bigl(\I + \mM^{-} \had (\diag(\vbeta)\,\mathbf{X}\mathbf{X}^{\T})\bigr)^{-1}\diag(\vbeta)$
for the WY matrix of \eqnref{eq:ut}.

\begin{proposition}[Chunk update]\label{prop:chunk}
Fix a chunk with keys $\bar{\mK}$ (rows $\bar{\vk}_r$), values $\mV$, chunk-initial states
$\ch{\Sone}{j}, \ch{\Stwo}{j}$, and step sizes $\vbeta^{(1)}, \vbeta^{(2)}$. Let $\mZ =
\bar{\mK}\,\ch{\Sone}{j}$, let $\bar{\mH}$ have rows $[\sigma(\vz_r); 1]$, let $\mE =
\bar{\mH}\,\ch{\Stwo}{j} - \mV$, and let $\mathbf{G}$ have rows $\bm{g}_r = (\ch{\Stwo}{j}\ve_r)
\had \sigma'(\vz_r)$ with the bias row dropped: the per-token errors of the two layers at the
chunk-initial states. Without forgetting and with the exact targets $\vtil_r = \vz_r -
\bm{g}_r$, \algref{alg:toaster} with the kernel on both layers (line~\ref{ln:wy}) returns
\begin{equation}\label{eq:chunk_closed}
  \ch{\Sone}{j+1} = \ch{\Sone}{j} - \bar{\mK}^{\T}\, \mT_{\bar{\mK}}\, \mathbf{G}, \qquad
  \ch{\Stwo}{j+1} = \ch{\Stwo}{j} - \bar{\mH}^{\T}\, \mT_{\bar{\mH}}\, \mE ,
\end{equation}
whereas one step of chunked gradient descent on $\sum_{r} \Loss_r$ at the chunk-initial
states, with per-token step sizes, gives the same expressions with $\mT_{\bar{\mK}}$ and
$\mT_{\bar{\mH}}$ replaced by $\diag(\vbeta^{(1)})$ and $\diag(\vbeta^{(2)})$.
\end{proposition}

\paragraph{Proof of Proposition~\ref{prop:chunk}.}
By \eqnref{eq:ut}--\eqnref{eq:chunk_alg}, the delta rule with keys $\mathbf{X}$ and values $\mU$
advances a state $\mS$ over the chunk to $\mS + \mathbf{X}^{\T}\mT_{\mathbf{X}}(\mU - \mathbf{X}\mS)$. Phase~II
applies it to $\ch{\Sone}{j}$ with keys $\bar{\mK}$ and values $\Vtil = \mZ - \mathbf{G} =
\bar{\mK}\,\ch{\Sone}{j} - \mathbf{G}$, so $\mU - \mathbf{X}\mS = -\mathbf{G}$; Phase~III applies it to
$\ch{\Stwo}{j}$ with keys $\bar{\mH}$ and values $\mV$, so $\mU - \mathbf{X}\mS = -\mE$. For
chunked gradient descent, the gradients of $\Loss_r$ at the chunk-initial states are
$\bar{\vk}_r \bm{g}_r^{\T}$ and $\bar{\vh}_r \ve_r^{\T}$ (proof of
Remark~\ref{prop:equiv}), and summing the $C$ steps gives
$-\bar{\mK}^{\T}\diag(\vbeta^{(1)})\mathbf{G}$ and $-\bar{\mH}^{\T}\diag(\vbeta^{(2)})\mE$. \qed

Both updates are therefore driven by the same per-token errors and differ only in the $C
\times C$ matrix that mixes them. Chunked gradient descent adds the $C$ steps, each computed
at the chunk-initial state, as in a Jacobi iteration; \algref{alg:toaster} applies them in
order, each residual re-evaluated against the running state, as in a Gauss--Seidel (Kaczmarz)
sweep toward the fixed targets $\Vtil$ and $\mV$. Since $\mathbf{L} = \mM^{-} \had
(\diag(\vbeta)\mathbf{X}\mathbf{X}^{\T})$ is strictly lower triangular, $\mT_{\mathbf{X}} - \diag(\vbeta) =
\sum_{m=1}^{C-1} (-\mathbf{L})^{m}\diag(\vbeta)$ exactly, so the two mixing matrices, and with them
the two updates for every residual matrix, coincide precisely when $\beta_r \bar{\vx}_r^{\T}\bar{\vx}_i = 0$
for all $i < r$ in the chunk (for a particular residual matrix the updates can also agree by
cancellation); with the homogeneous coordinate, $\bar{\vk}_i^{\T}\bar{\vk}_r = \vk_i^{\T}\vk_r + 1$,
orthogonal keys no longer satisfy this condition: the bias coordinate adds $1$ to every
pairwise inner product in the chunk. Neither update is closer to
per-token gradient descent in general: within a chunk, \algref{alg:toaster} fits fixed
targets, whereas chunked gradient descent adds fixed gradients, and the errors $\mathbf{G}$ are
equally stale in both. With apply-then-update on the hidden layer (line~\ref{ln:atu}, our
default) the two forms are mixed: the hidden update is the chunked-gradient expression
$\ch{\Sone}{j} - \bar{\mK}^{\T}\diag(\vbeta^{(1)})\mathbf{G}$, a Jacobi step toward the fixed
targets, and the output update keeps $\mT_{\bar{\mH}}$, the Gauss--Seidel sweep.

\paragraph{The practical target is a descent direction.}
With the bounded target of \eqnref{eq:target_practical}, the hidden update becomes
$-\beta^{(1)}_t \vk_t \hat{\bm{g}}_t^{\T}$ with $\hat{\bm{g}}_t = \Norm(\Stwo_{t-1}\ve_t)
\had \sigma'(\vz_t)$, whereas the gradient direction is $\bm{g}_t = (\Stwo_{t-1}\ve_t) \had
\sigma'(\vz_t)$. Since $\Norm(x) = \gamma\,\softsign(x)$ has the sign of $x$ for every
coordinate, $\langle \bm{g}_t, \hat{\bm{g}}_t \rangle = \sum_i (\Stwo_{t-1}\ve_t)_i\,
\Norm\bigl((\Stwo_{t-1}\ve_t)_i\bigr)\, \sigma'(\vz_t)_i^2 \ge 0$, and therefore
$\langle \nabla_{\Sone}\Loss_t,\; \vk_t \hat{\bm{g}}_t^{\T} \rangle = \lVert \vk_t
\rVert^2 \langle \bm{g}_t, \hat{\bm{g}}_t \rangle \ge 0$: the update is a descent
direction for $\Loss_t$, although no longer the gradient. The argument assumes $\gamma > 0$, no
forgetting, and a residual $\ve_t$ evaluated at the states $(\Sone_{t-1}, \Stwo_{t-1})$ at
which the gradient is taken. It concerns the direction of one per-token step only: it does
not bound the loss after the finite step $\beta^{(1)}_t$ and gives no monotone-loss or
stability guarantee. The default chunked algorithm (line~\ref{ln:atu} of \algref{alg:toaster})
evaluates its residuals at the chunk-initial states and applies the chunk-averaged forget
gate, so it lies outside this argument.

\section{The WY representation and the chunkwise kernel}
\label{app:wy}

This appendix derives the chunkwise algorithm behind \kernel{} for the ungated delta rule,
following \citet{yang2024parallelizing}, in the state convention $\mS_t \in \R^{\dk \times
\dv}$, $\vo_t = \mS_t^{\T}\vq_t$ used throughout the paper. The gated variants used in our
experiments follow the same pattern and are described at the end.

\paragraph{Unrolling a chunk.}
Consider chunk $j$ with tokens $r = 1, \dots, C$, chunk-initial state $\mS^{0} = \ch{\mS}{j}$,
and $\mS^{r}$ the state after token $r$. Unrolling $\mS^{r} = (\I - \beta_r \vk_r
\vk_r^{\T})\mS^{r-1} + \beta_r \vk_r \vv_r^{\T}$ gives
\begin{equation}\label{eq:unroll}
  \mS^{r} = \mP^{r}\, \mS^{0} + \mH^{r}, \qquad
  \mP^{r} = \prod_{s=r}^{1} (\I - \beta_s \vk_s \vk_s^{\T}), \qquad
  \mH^{r} = \sum_{i=1}^{r} \Bigl[\prod_{s=r}^{i+1} (\I - \beta_s \vk_s \vk_s^{\T})\Bigr] \beta_i \vk_i \vv_i^{\T},
\end{equation}
where products are ordered with the largest index on the left.

\paragraph{WY form.}
Products of generalized Householder matrices admit the compact
representation~\citep{bischof1987wy}
\begin{equation}\label{eq:wy}
  \mP^{r} = \I - \sum_{i=1}^{r} \vk_i \vw_i^{\T}, \qquad
  \mH^{r} = \sum_{i=1}^{r} \vk_i \vu_i^{\T},
\end{equation}
with auxiliary vectors defined by the recursions
\begin{equation}\label{eq:wy_rec}
  \vw_r = \beta_r \Bigl(\vk_r - \sum_{i<r} (\vk_r^{\T}\vk_i)\, \vw_i\Bigr), \qquad
  \vu_r = \beta_r \Bigl(\vv_r - \sum_{i<r} (\vk_r^{\T}\vk_i)\, \vu_i\Bigr).
\end{equation}
Both follow by induction from $\mP^{r} = (\I - \beta_r\vk_r\vk_r^{\T})\mP^{r-1}$ and
$\mH^{r} = (\I - \beta_r\vk_r\vk_r^{\T})\mH^{r-1} + \beta_r\vk_r\vv_r^{\T}$. Stacking the
auxiliary vectors as rows of $\mW \in \R^{C\times\dk}$ and $\mU \in \R^{C\times\dv}$, the
recursions \eqnref{eq:wy_rec} become a single triangular system (the UT
transform~\citep{joffrain2006accumulating}),
\begin{equation}\label{eq:ut}
  \mW = \mT\,\mK, \qquad \mU = \mT\,\mV, \qquad
  \mT = \Bigl(\I + \mM^{-} \had \bigl(\diag(\vbeta)\,\mK\mK^{\T}\bigr)\Bigr)^{-1} \diag(\vbeta),
\end{equation}
where $\mM^{-}$ is the strictly lower-triangular mask and the inverse is a $C \times C$
forward substitution.

\paragraph{Chunk state and outputs.}
Substituting \eqnref{eq:wy} into \eqnref{eq:unroll} at $r = C$ and reading out
$\vo_r = (\mS^{r})^{\T}\vq_r$ for every $r$ gives
\begin{equation}\label{eq:chunk_alg}
  \ch{\mS}{j+1} = \ch{\mS}{j} + \mK^{\T}\bigl(\mU - \mW\,\ch{\mS}{j}\bigr), \qquad
  \mO = \mQ\,\ch{\mS}{j} + \bigl(\mQ\mK^{\T} \had \mM\bigr)\bigl(\mU - \mW\,\ch{\mS}{j}\bigr),
\end{equation}
with $\mM$ the lower-triangular mask including the diagonal. Every operation in
\eqnref{eq:ut}--\eqnref{eq:chunk_alg} is a dense matrix product (or a small triangular
solve), the state is propagated sequentially only across chunks, and the result is exact
for any chunk size $C$. This is the kernel \kernel{} of \eqnref{eq:kernel}; \model calls
it once or twice per chunk with the arguments listed in \algref{alg:toaster}.

\paragraph{Gated variants.}
A scalar forget gate $\alpha_t$ (Gated DeltaNet) commutes with the Householder factors, so
the cumulative decay $\gamma^{r} = \prod_{s \le r}\alpha_s$ can be pulled out of
$\mP^{r}$ and absorbed into decay-weighted keys and a decay-aware causal mask in
\eqnref{eq:ut}--\eqnref{eq:chunk_alg}~\citep{yang2025gated}. A channel-wise gate
$\Diag(\valpha_t)$ (KDA) does not commute, and the cumulative diagonal decays instead enter
the inner products $\vk_r^{\T}\vk_i$ of \eqnref{eq:wy_rec} and the corresponding entries of
$\mT$~\citep{team2025kimi}. In both cases the algorithm keeps the structure of
\eqnref{eq:chunk_alg}, and \model uses the KDA form of the kernel unchanged for every layer
it advances with the kernel.

\section{Operation count}
\label{app:cost}

We count multiply--accumulate operations per token and per head in the forward pass,
ignoring the $O(C^3)$ triangular solve and pointwise work. For one call of \kernel{} with
input width $d_{\mathrm{in}}$, output width $d_{\mathrm{out}}$, and chunk size $C$,
\eqnref{eq:ut}--\eqnref{eq:chunk_alg} cost approximately
\begin{equation}
  3\, d_{\mathrm{in}} d_{\mathrm{out}} \;+\; 3\, C\, d_{\mathrm{in}} \;+\; 2\, C\, d_{\mathrm{out}}
\end{equation}
per token (three products involving the state and five intra-chunk products of size $C$).
KDA at $\dk = \dv = 64$, $C = 32$ therefore costs about $2.3 \times 10^{4}$ operations per
token and head. \model makes two calls, $\dk \to \dhid$ and $\dhid \to \dv$, and adds the
three Phase~I products $\ch{\mK}{j}\ch{\Sone}{j}$, $\ch{\mH}{j}\ch{\Stwo}{j}$, and
$\ch{\mE}{j}\ch{\Stwo}{j}{}^{\T}$, each $\dk\dhid$ (or $\dhid\dv$) per token; with $\dhid =
256$ the total is about $2.0 \times 10^{5}$ operations per token and head, roughly nine
times KDA. With the default apply-then-update hidden layer (line~\ref{ln:atu} of
\algref{alg:toaster}) the first call is replaced by the two products $\ch{\mQ}{j}\ch{\Sone}{j}$
and $\ch{\mK}{j}{}^{\T}\Diag(\vbeta^{(1)})\mathbf{G}_{[j]}$, $2\dk\dhid$ per token, and
the total falls to about $1.6 \times 10^{5}$, roughly seven times KDA. For comparison, the dense projections of a block at width $D = 2048$ (query,
key, value, and output projections plus a SwiGLU feed-forward network of inner width $4D$)
cost about $16 D^2 \approx 6.7 \times 10^{7}$ operations per token. With $32$ heads, the
memory kernels thus account for roughly $1\%$ of a block's operations for KDA and roughly
$9\%$ for \model with two kernel calls ($8\%$ with the default) at the 1.2B scale, and proportionally less at larger widths, since the
kernel cost grows with the number of heads while the dense cost grows with $D^2$. At 145M
($D = 1024$, $16$ heads) the shares are about $2\%$ for KDA and $15$--$19\%$ for \model. These are
forward-pass operation counts (the backward pass scales all terms alike), excluding the
$O(C^3)$ triangular solves and the gate projections; they are not measured throughput.
Chunkwise kernels at $C = 32$ are often bound by state traffic rather than arithmetic, and
\model reads and writes about eight times KDA's state per chunk, with Phase~III dependent on
Phase~II; we have not measured utilization, and the larger state traffic may lower it. The
chunk loop also remains sequential: a chunk's hidden keys and targets depend on the
chunk-initial states, so context parallelism across chunks does not apply.

\paragraph{State size.}
Per head, KDA stores $\dk\dv$ numbers and \model stores $(\dk+1)\dhid + (\dhid+1)\dv$; with
$\dk = \dv = 64$ and $\dhid = 256$ this is $33{,}088$ numbers, about eight times KDA. At the
1.2B scale (24 layers, 32 heads) the full \model state is $2.5 \times 10^{7}$ numbers per
sequence, constant in the sequence length; a softmax-attention key--value cache at the same
scale holds $2 \cdot 24 \cdot (32 \cdot 64) = 9.8 \times 10^{4}$ numbers \emph{per token} in the same precision, i.e., it exceeds the \model state beyond a context
of about 260 tokens.

\section{Experimental details}
\label{app:hparams}

\subsection{Scaling ladder}
\label{app:ladder}

\begin{table}[h]
  \caption{Scaling ladder. Depth, width, and number of heads grow together; the head
  dimension is 64, the feed-forward inner width is $4D$, and the context length is 4,096
  tokens throughout. Parameter counts are those reported for the ladder configuration; the
  recurrent mixers add a small number of parameters for their gates and convolutions. The
  token budget is roughly proportional to model size (46 to 40 tokens per non-embedding
  parameter).}
  \label{tab:ladder_config}
  \centering
  \footnotesize
  \setlength{\tabcolsep}{3.2pt}
  \begin{tabular}{@{}l rrr rr rrr rr@{}}
    \toprule
    Size & Layers & Width $D$ & Heads & Non-emb. & Total & Batch & Steps & Tokens & Peak LR & Weight decay \\
    \midrule
    145M & 12 & 1024 & 16 & 201M & 234M & 64  & 35,050 & 9.2B  & $1.95\times10^{-3}$ & $2.28\times10^{-4}$ \\
    300M & 15 & 1280 & 20 & 390M & 431M & 96  & 42,708 & 16.8B & $1.56\times10^{-3}$ & $1.87\times10^{-4}$ \\
    550M & 18 & 1536 & 24 & 671M & 721M & 160 & 42,630 & 27.9B & $1.30\times10^{-3}$ & $1.88\times10^{-4}$ \\
    850M & 21 & 1792 & 28 & 1.06B & 1.12B & 192 & 55,095 & 43.3B & $1.12\times10^{-3}$ & $1.45\times10^{-4}$ \\
    1.2B & 24 & 2048 & 32 & 1.58B & 1.65B & 256 & 60,725 & 63.7B & $9.77\times10^{-4}$ & $1.32\times10^{-4}$ \\
    \bottomrule
  \end{tabular}
\end{table}

\paragraph{Sizes and token budgets.}
Every architecture is trained at the five sizes of \tabref{tab:ladder_config}, from 12
layers of width 1024 to 24 layers of width 2048, scaling depth, width, and number of heads
together. We refer to the sizes by the nominal labels 145M, 300M, 550M, 850M, and 1.2B; their
non-embedding parameter counts are 0.20B, 0.39B, 0.67B, 1.06B, and 1.58B. The token budget,
batch size, and peak learning rate are scaled with model size along a single scaling rule
(9.2B to 63.7B tokens, i.e., 46 to 40 tokens per non-embedding parameter;
\tabref{tab:ladder_config}), so that every point of the ladder is trained under the same
regime. At a given size all architectures share every backbone, data, and optimizer
setting; only the sequence mixer changes.

\paragraph{Data and tokenization.}
All models are pre-trained on C4~\citep{raffel2020exploring} (version 3.1.0). Documents are
concatenated and packed into sequences of 4,096 tokens, with beginning- and end-of-sequence
tokens marking document boundaries. We use a unigram SentencePiece
tokenizer~\citep{kudo2018sentencepiece} with a vocabulary of 32,000 subwords plus 100
reserved tokens, shared by all architectures together with the embedding matrix. Validation
loss is measured on the C4 validation split (in-domain) and, out of domain, on
PG19~\citep{rae2020compressive}, full-length books that stress long-range coherence, each at
the 4,096-token context. Perplexity is $\exp$ of the mean per-token loss.

\paragraph{Optimization.}
All models use AdamW~\citep{loshchilov2019decoupled} with moment coefficients $(0.9,
0.95)$ and decoupled weight decay applied to weight matrices only (not to biases or
normalization scales). The learning rate warms up linearly from zero over the first 1,000
steps to the peak in \tabref{tab:ladder_config} and decays linearly to zero at the last
step; the global gradient norm is clipped to $1.0$. Master weights, gradients, and the
recurrent memory states of all recurrent models are kept in 32-bit floating point. All
numbers in the paper are single training runs; we lacked the compute to repeat them and
therefore rest the evidence on the consistency of the ordering across sizes and corpora.

\subsection{Training compute, scaling fits, and compute-efficiency gain}
\label{app:scaling_fit}

\paragraph{Training compute.}
We measure the training compute of every ladder point as $C = 6ND$ FLOPs, where $N$ is the
number of non-embedding parameters of the ladder configuration (\tabref{tab:ladder_config})
and $D$ the number of training tokens: the forward pass costs about $2N$ FLOPs per token and
the backward pass about $4N$~\citep{kaplan2020scaling, hoffmann2022training}. The five points
of the ladder correspond to $1.1 \times 10^{19}$, $3.9 \times 10^{19}$, $1.1 \times 10^{20}$,
$2.8 \times 10^{20}$, and $6.0 \times 10^{20}$ FLOPs. The approximation ignores the
sequence-length-dependent cost of attention (quadratic in the 4,096-token context) and the
memory kernels of the recurrent models (about $1\%$ of a block's multiply--accumulates for KDA
and $8$--$9\%$ for \model at 1.2B; \appref{app:cost}), and it uses the same $N$ for all three
architectures, whose mixers differ only by a small number of gate and convolution parameters
and, for \model, the learned initial states, which add no per-token operations.
The three models at a given size therefore share one compute budget, and the comparisons
below are comparisons of loss at matched $6ND$ compute, not of wall-clock time
(\figref{fig:throughput}).

\paragraph{Scaling fits.}
For each architecture and corpus we fit the power law $L(C) = (C_c / C)^{\alpha}$, i.e.,
$\log L = a - \alpha \log C$, by least squares in log--log space to the five points of the
ladder, the form used by \citet{kaplan2020scaling} for the compute frontier.
\tabref{tab:scaling_fits} lists the fitted exponents. Every fit has $R^2 \ge 0.9996$ and a
maximum residual of $0.006$ nats, and the exponents
($0.045$--$0.047$) are close to the frontier exponent of $0.050$ reported by
\citet{kaplan2020scaling}. A fit with an irreducible term, $L(C) = E + A\,C^{-\alpha}$
\citep{hoffmann2022training}, is not identifiable from five points: on PG19 the fitted $E$ is
zero and on C4 it trades off against $\alpha$ ($E = 0.9$, $\alpha = 0.07$). We therefore
report the two-parameter fit; the compute-efficiency gains below change by less than one
percentage point under the three-parameter form, because they depend only on the local slope
of Attention's curve.

\begin{figure}[H]
  \centering
  \includegraphics[width=0.84\textwidth]{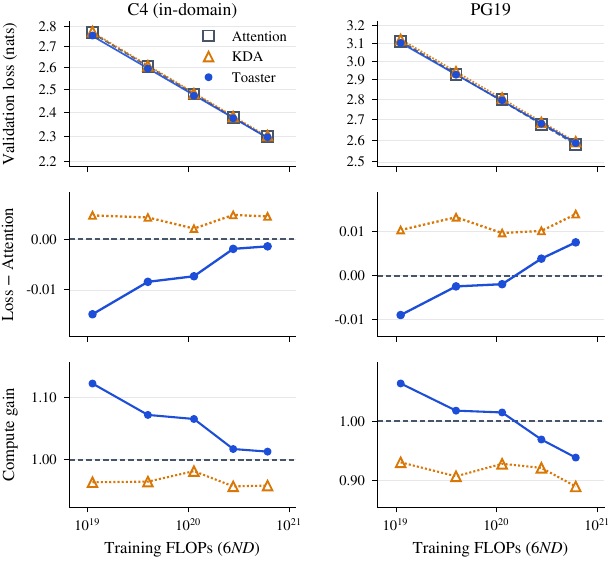}
  \caption{\textbf{Scaling ladder on C4 and PG19} (single runs). Top: validation loss
  against training compute ($6ND$ FLOPs) with the power-law fits of \appref{app:scaling_fit}.
  Middle: loss minus Attention's loss at the same compute (below zero is
  better than Attention). Bottom: compute-efficiency gain relative to Attention,
  \eqnref{eq:ceg}, at the measured sizes.}
  \label{fig:scaling_full}
\end{figure}

\paragraph{Compute-efficiency gain.}
We convert a loss difference at matched compute into a compute ratio through Attention's
fitted law. If Attention's loss falls as $C^{-\alpha_{\mathrm{A}}}$, lowering it from
$L_{\mathrm{A}}$ to a model's loss $L_{\mathrm{m}}$ at the same compute requires multiplying
Attention's compute by
\begin{equation}\label{eq:ceg}
  g \;=\; \bigl(L_{\mathrm{A}} / L_{\mathrm{m}}\bigr)^{1/\alpha_{\mathrm{A}}} ,
\end{equation}
which we call the compute-efficiency gain of the model relative to Attention: $g > 1$ means
that Attention needs $g$ times the model's compute to reach the model's loss, $g < 1$ that it
needs less. The
gain panels of \figref{fig:scaling} and \figref{fig:scaling_full} show $g$ at the five
measured sizes, joined by straight segments; no curve is fitted to the gains, and the fitted
loss laws elsewhere are drawn over the measured compute range only, without extrapolation.
Because $\alpha_{\mathrm{A}}
\approx 0.046$, a loss difference of $0.01$ nats (about $0.4\%$) corresponds to about $9\%$ of
compute, which is why the gain separates the architectures more visibly than the loss or
perplexity curves.

\begin{table}[h]
  \caption{Power-law fits $L = (C_c/C)^{\alpha}$ of validation loss against training compute
  (log--log least squares over the five sizes) and the compute-efficiency gain
  relative to Attention of \eqnref{eq:ceg} at the five measured sizes ($g > 1$: Attention
  needs $g$ times the model's compute to reach the model's loss).}
  \label{tab:scaling_fits}
  \centering
  \small
  \setlength{\tabcolsep}{5pt}
  \begin{tabular}{@{}l l cc ccccc@{}}
    \toprule
    & & & & \multicolumn{5}{c}{Compute-efficiency gain vs.\ Attention} \\
    \cmidrule(lr){5-9}
    Corpus & Model & $\alpha$ & $R^2$ & 145M & 300M & 550M & 850M & 1.2B \\
    \midrule
    \ladderfitrows
    \bottomrule
  \end{tabular}
\end{table}

\paragraph{Why $6ND$, and what the sequence mixers cost.}
We report compute as $6ND$ rather than as a count that includes each sequence mixer's own
operations, for three reasons. First, it is the convention of compute scaling laws:
\citet{kaplan2020scaling} likewise leave the context-dependent cost of attention out of $C$,
since it is a small fraction of the total when $D \gg n_{\mathrm{ctx}}/12$ (on our ladder $D$
is three to six times $n_{\mathrm{ctx}}/12$). Second, $6ND$ counts what the three
architectures share at each size---the backbone, $N$, and the training tokens---so that the
gain of \eqnref{eq:ceg} measures what the mixer contributes per parameter and per training
token. Third, and most important for scaling, at a fixed context the mixer's own cost becomes
a small part of the total as the model grows. The dense projections of a block cost $16D^2$
operations per token. A recurrent mixer costs a fixed number of operations per head and
token, set by its head dimensions and chunk size and independent of $D$ (\appref{app:cost}),
and causal attention costs about $TD$ at context $T$. Relative to the dense projections, the
mixers therefore shrink as $1/D$: about $22/D$ for KDA, $156/D$ for \model ($194/D$ with the
kernel on both layers), and $T/16D = 256/D$ for attention at $T = 4{,}096$. The top block of
\tabref{tab:mixer_cost} lists these shares along the ladder; at width $4{,}096$ they would be
about $0.5\%$, $4\%$ ($5\%$), and $6\%$.

\paragraph{Gains with the mixers counted.}
Counting the mixers multiplies each model's compute by $1+o$, where $o$ is its share in
\tabref{tab:mixer_cost} (the backward pass scales all terms alike; gate projections and
triangular solves are left out, as in \appref{app:cost}), and turns the gain of
\eqnref{eq:ceg} into $g\,(1+o_{\mathrm{A}})/(1+o_{\mathrm{m}})$; refitting Attention's law on
the mixer-inclusive compute changes these values by less than half a percentage point.
Against Attention, whose quadratic term is the largest mixer cost at 4K, every gain rises:
\model's C4 gain becomes 1.22 at 145M and 1.06 at 1.2B, and KDA's 1.18 and 1.07. Against
KDA, the extra kernel arithmetic of \model offsets its gain at these sizes: the ratio of
\model's gain to KDA's is 1.06--1.16 on C4 and 1.05--1.14 on PG19 at matched $6ND$, and
0.99--1.03 and 0.98--1.02 at matched mixer-inclusive compute. On our ladder \model's
advantage over KDA is therefore one per parameter and per training token. Its extra cost and
this advantage both shrink with scale (on C4, from 13\% to 6\% and from 16\% to 6\% of
compute), and whether the advantage outlasts the cost at larger widths is untested. These
are operation counts, not measured time; \figref{fig:throughput} reports step time.

\begin{table}[h]
  \caption{Sequence-mixer cost and the compute-efficiency gain with it counted. Top: forward
  operations of the sequence mixer per token as a share of a block's dense projections
  ($16D^2$; \appref{app:cost}), for Attention at the 4,096-token context. Middle: C4 gain
  relative to Attention, \eqnref{eq:ceg}, with each model's compute multiplied by one plus its
  share. Bottom: ratio of \model's gain to KDA's at matched $6ND$ and at matched mixer-inclusive
  compute.}
  \label{tab:mixer_cost}
  \centering
  \small
  \setlength{\tabcolsep}{5pt}
  \begin{tabular}{@{}l l ccccc@{}}
    \toprule
    & & 145M & 300M & 550M & 850M & 1.2B \\
    \midrule
    Mixer share & Attention & 25.0\% & 20.0\% & 16.7\% & 14.3\% & 12.5\% \\
    & KDA & 2.1\% & 1.7\% & 1.4\% & 1.2\% & 1.1\% \\
    & \model & 15.2\% & 12.2\% & 10.2\% & 8.7\% & 7.6\% \\
    & \model, kernel on both layers & 18.9\% & 15.2\% & 12.6\% & 10.8\% & 9.5\% \\
    \midrule
    C4 gain, mixer-inclusive & KDA & 1.18 & 1.14 & 1.13 & 1.08 & 1.07 \\
    & \model & 1.22 & 1.15 & 1.13 & 1.07 & 1.06 \\
    \midrule
    \model{} vs.\ KDA, C4 & matched $6ND$ & 1.16 & 1.11 & 1.08 & 1.06 & 1.06 \\
    & mixer-inclusive & 1.03 & 1.01 & 1.00 & 0.99 & 0.99 \\
    \model{} vs.\ KDA, PG19 & matched $6ND$ & 1.14 & 1.12 & 1.09 & 1.05 & 1.05 \\
    & mixer-inclusive & 1.01 & 1.02 & 1.01 & 0.98 & 0.99 \\
    \bottomrule
  \end{tabular}
\end{table}

\subsection{Shared backbone}
\label{app:backbone}

All models are built on NanoDO~\citep{liu2024nanodo}, a minimal decoder-only Transformer
implementation in JAX, and differ only in the sequence mixer: the attention block of every
layer (of three layers in four in the hybrids of \tabref{tab:downstream}) is replaced by the mixer
under comparison, and everything else---embeddings,
normalization, feed-forward blocks, output head, data pipeline, and optimizer---is shared
unchanged. Every model is a pre-normalized decoder: RMSNorm~\citep{zhang2019root} ($\epsilon =
10^{-6}$) precedes the sequence mixer and the feed-forward network of each block, and a
final RMSNorm precedes the output projection. The feed-forward network is a
SwiGLU~\citep{shazeer2020glu} with inner width $4D$. In the attention baseline, queries and keys carry rotary position embeddings~\citep{su2024roformer} with base $10{,}000$ over the head dimension of $64$; the recurrent mixers use no positional encoding (\secref{sec:practical}).
Linear layers are initialized with Xavier-uniform weights and embeddings from a zero-mean
normal distribution scaled by the fan-in. The recurrent mixers apply a depthwise causal
convolution of width 4 to the projected queries, keys, and values before the memory, as in
KDA.

\subsection{Sequence mixers}
\label{app:mixers}

\paragraph{Attention.}
Causal multi-head softmax attention with 64-dimensional heads and RMSNorm applied to queries
and keys before the dot product.

\paragraph{KDA.}
The chunkwise gated delta rule of \citet{team2025kimi} with chunk size $C = 32$,
$\ell_2$-normalized keys and queries, input-dependent channel-wise forget gates and step
sizes, a step-size scale of $1.0$, and a decay scale of $0.1$.

\paragraph{Titans.}
A deep test-time-training memory~\citep{behrouz2026titans} whose two-layer MLP has hidden
width $\dhid = \dk = 64$, updated by gradient descent on the reconstruction loss with the stale
chunkwise gradient of \eqnref{eq:stale}, momentum, and weight decay. It is timed at chunk sizes
$C = 16$ and $C = 32$ in the throughput comparison (\appref{app:throughput}) and trained at 1.2B,
with all layers replaced and in the hybrid, at the same hidden width $\dhid = 64$ and $C = 32$
(\tabref{tab:downstream}).
Our Titans implementation builds on the public JAX code base of TTT~\citep{sun2024learning}.

\paragraph{\model.}
The configuration used in every main-text experiment except the throughput profile, which
uses a smaller hidden width, matched to Titans' memory (\appref{app:throughput}): two KDA memories in cascade with
hidden width $\dhid = 4\dk = 256$, GELU activation,
homogeneous-coordinate bias (\secref{sec:practical}), the hidden memory advanced by
apply-then-update (line~\ref{ln:atu} of \algref{alg:toaster}, with the chunk-averaged forget
gate) and the output memory by the KDA kernel, RMSNorm on the value stream, the
bounded back-projected error of \eqnref{eq:target_practical} with $\gamma = 0.01$,
input-dependent channel-wise forget gates and step sizes produced by rank-64 projections of
the input, a step-size scale of $0.13$, a decay scale of $0.01$, and chunk size $C = 32$.
Unlike KDA, \model does not $\ell_2$-normalize the keys $\bar{\vh}_t$ or the queries of the
output memory; nothing therefore bounds $\beta^{(2)}_t \lVert \bar{\vh}_t \rVert^2$ by
construction, but we have not found this to be an issue in practice.
The memory states are accumulated in 32-bit floating point, as for KDA. The state per head
is $(\dk + 1)\dhid + (\dhid + 1)\dv = 33{,}088$ numbers, about eight times KDA's $\dk\dv =
4{,}096$ (\appref{app:cost}).

\paragraph{Hybrid models.}
The hybrids of \tabref{tab:downstream} keep the 1.2B backbone of \tabref{tab:ladder_config} and
use the attention block of the Attention baseline (rotary embeddings, 64-dimensional heads) in
every fourth layer (layers 4, 8, \ldots, 24); the other 18 layers use the mixer under comparison
with the settings above. SWA is this attention block restricted to a causal window of the 128
most recent tokens and without rotary embeddings; in every hybrid, only the six global attention
layers carry a positional encoding. Data, token budget, and optimizer settings are those of the
1.2B point of the ladder.

\subsection{Downstream evaluation}
\label{app:eval}

All downstream tasks are evaluated three-shot on their standard evaluation splits (the
validation split where labeled, e.g., the SuperGLUE tasks): three in-context
demonstrations of the task precede each query, and there is no task-specific fine-tuning
or prompt engineering. For multiple-choice and classification tasks, each candidate
completion is scored by its length-normalized conditional log-likelihood given the prompt
and the highest-scoring candidate is selected. Evaluation runs use a batch size of 256 and
the full 4,096-token context, data-parallel over four accelerators. The 13 tasks, grouped
into the six skill categories of \tabref{tab:downstream_full}, are ARC-Easy and
ARC-Challenge~\citep{clark2018think} (science QA); COPA~\citep{roemmele2011choice},
HellaSwag~\citep{zellers2019hellaswag}, OpenBookQA~\citep{mihaylov2018can},
PIQA~\citep{bisk2020piqa}, and SWAG~\citep{zellers2018swag} (commonsense reasoning);
MultiRC~\citep{khashabi2018looking} (reading comprehension); MathQA~\citep{amini2019mathqa}
(mathematical reasoning); CB~\citep{demarneffe2019commitmentbank},
MNLI~\citep{williams2018broad}, and RTE~\citep{wang2019glue, dagan2005pascal} (NLI and
entailment); and SST-2~\citep{socher2013recursive} (linguistic understanding), several of
them through SuperGLUE~\citep{wang2019superglue}. Six of the tasks (SST-2, MultiRC, MNLI,
MathQA, CB, RTE) are at or near the chance or majority-class baseline for every model at
this scale; on the remaining seven (ARC-Easy, ARC-Challenge, COPA, HellaSwag, OpenBookQA,
PIQA, SWAG) the averages are 56.5, 57.1, 56.1, and 57.7 for Attention, KDA, Titans, and
\model with all layers replaced, and 57.8, 57.2, 56.7, and 58.3 for the SWA, KDA, Titans, and
\model hybrids. Among Attention, KDA, and \model, the largest per-task margins (2--7 points on
COPA, CB, and RTE) fall on the smallest evaluation sets (100, 56, and 277 items, with binomial
standard errors of 3--7 points), and no difference on an evaluation set of more than a thousand
items exceeds two points; Titans with all layers replaced is the exception, 2--6 points below
\model on ARC-Easy, HellaSwag, and PIQA, and its hybrid is 4 points below \model's on ARC-Easy.

\subsection{Length extrapolation}
\label{app:extrapolation}

\figref{fig:extrapolation} evaluates the 1.2B models, trained with a 4,096-token context,
on sequences five times longer than anything seen in training. We take the novels of the
English long-book question-answering task (En.QA) of
$\infty$Bench~\citep{zhang2024infinitebench}, in which the names of key entities are
replaced throughout each book to prevent contamination, and draw one sequence of 20,480
tokens from each of 100 books. Each model scores every sequence with its final
checkpoint (60,725 steps), and we record the per-token loss at every position, averaged over
the 100 sequences; the perplexity at a position is $\exp$ of that mean loss. The curves in
\figref{fig:extrapolation} are a one-sided running average of the per-position mean loss
over the preceding positions, which is why the first thousand positions reflect the warm-up
of the average; the shaded region marks positions beyond the training context, and the
perplexity axis is capped at 30, which Attention exceeds near position 10K. Attention is evaluated with the rotary base it was trained with and no
position-extrapolation technique; the recurrent models need none, since their state does not
grow with the context. \tabref{tab:extrapolation_windows} reports the mean loss of
each model in 4K-token windows: \model's is below KDA's by $0.008$--$0.014$ nats in every
window and by $0.009$ nats in the last (16--20K); beyond the first thousand positions, the
running-average loss of \model is also below KDA's at every position.

\subsection{Needle in a haystack}
\label{app:niah}

We use the three single-needle tasks of RULER~\citep{hsieh2024ruler}, in which a key--value
pair (the needle) is inserted into a long context (the haystack) and the model is asked,
after the context, to return the value associated with the key. S-NIAH-1 uses a passkey (a
number) as the needle and repeated synthetic sentences as the haystack; S-NIAH-2 uses a
number as the needle and essays as the haystack; S-NIAH-3 uses a UUID as the needle and the
same essay haystack. We keep RULER's task definitions and prompt template but regenerate the
haystacks for our setting: each prompt is built to fill exactly the target context length
of 512, 1K, 2K, or 4K tokens under our tokenizer (\appref{app:ladder}), and every generated
prompt is verified to fit its length budget with the query intact. The needle is inserted at
15 depths spaced evenly from the start (0\%, farthest from the query) to the end (100\%)
of the haystack. Accuracy is the fraction of examples per cell for which the
value is retrieved exactly, with greedy decoding and no instruction tuning, using the final
1.2B checkpoints; \tabref{tab:niah} averages it over the 15 depths, and
Figures~\ref{fig:niah_s1}--\ref{fig:niah_s3} report every cell.

\subsection{Throughput measurement}
\label{app:throughput}

\figref{fig:throughput} reports the wall-clock time of one training step (forward and
backward pass with the optimizer update) for context lengths from 2K to 32K tokens, with
the batch size reduced as the context grows so that every step processes 512K tokens.
Each configuration is timed over three runs with different seeds and the mean is reported;
the standard deviations are small compared with the differences between architectures. The
measurements use the 145M configuration on eight TPU~v6 chips (a $2 \times 2
\times 2$ slice). Attention
uses Flash Attention, and KDA and \model use chunk size $C = 32$; Titans is timed at $C = 16$
and $C = 32$, since its stale chunkwise update becomes less accurate with $C$.
The \model timed here is not the default configuration of our quality experiments.
For a fair comparison of training throughput we match the fast-weight dimensions of the two deep memories:
\model is timed at hidden width $\dhid = \dk = 64$, the hidden width of the Titans memory
MLP, instead of the default $\dhid = 4\dk = 256$ (\appref{app:mixers}), so that \model and
Titans have fast weights of the same width per head (\model's bias adds one row to each
matrix; Titans' momentum buffer, which has the size of its weights, is not counted). We call this configuration width-matched. All three recurrent models use 64-dimensional
heads. Only the update rules are those of the default: apply-then-update on the hidden memory
and the KDA kernel on the output memory. We have not optimized the \model implementation for
speed: it is written naively in JAX, calling the KDA chunkwise routine unchanged for the
output memory and computing the hidden-memory update and the cross-layer signals with plain
dense operations, with no fused or hardware-specific kernel. Improving its speed, for example
with a fused kernel for the cascade, is left to future work. At 32K, one step
takes 1,085 ms for KDA, 1,138 ms for Flash Attention, 1,444 ms for \model, and 4,097 and
7,980 ms for Titans at $C = 32$ and $C = 16$.

\paragraph{Default configuration.}
\figref{fig:throughput} times \model at the matched width $\dhid = 64$. To connect the
throughput to the quality results, we also time the default configuration used in every
other experiment ($\dhid = 4\dk = 256$; \appref{app:mixers}), with the same update rules and
the same protocol; \tabref{tab:throughput_default} lists the step times of all
configurations. Widening the hidden memory four-fold, which about quadruples the state per
head (\appref{app:ablation}), increases the step time by only 7--20\% over the width-matched
\model. The default configuration takes $1.25$--$1.48\times$ the step time of KDA
($1.42\times$ at 32K) and $1.36$--$1.73\times$ that of Flash Attention, and its step time
grows by $1.40\times$ from 2K to 32K, against $1.23\times$ for KDA and $3.4$--$4.5\times$ for
Titans. Although its fast weights are about four times those of the Titans memory, it is faster than
Titans at every context length except 4K, where Titans at $C = 32$ is $0.7\%$ faster (1,215
against 1,223 ms); at 32K it is $2.7\times$ and $5.2\times$ faster than Titans at $C = 32$
and $C = 16$.

Most of the overhead over KDA is thus present already at $\dhid = 64$ ($1.12$--$1.33\times$
KDA) and does not come from the wider hidden memory. It is also larger than the operation
counts suggest: at 145M the memory kernels account for about $15\%$ of a block's operations
for the default \model and $2\%$ for KDA (\tabref{tab:mixer_cost}). We expect the remainder
to come from the unfused implementation described above and from the larger state traffic
noted in \appref{app:cost}, but we have not profiled these sources separately. At 1.2B the
kernels' share falls to about $8\%$ for \model and $1\%$ for KDA, so we expect a smaller
relative overhead at the scale of our main results; we have timed only the 145M
configuration.

\begin{table}[h]
  \caption{Training-step time (ms) at a fixed budget of 512K tokens per step, measured on the
  145M configuration. The first \model row is the width-matched
  configuration of \figref{fig:throughput}; the second is the default configuration of all
  quality experiments. Both advance the hidden memory by apply-then-update and the output
  memory by the KDA kernel. The last two rows give ratios of step times.}
  \label{tab:throughput_default}
  \centering
  \small
  \setlength{\tabcolsep}{5pt}
  \begin{tabular}{@{}l l ccccc@{}}
    \toprule
    Model & Setting & 2K & 4K & 8K & 16K & 32K \\
    \midrule
    Flash Attention &  & 639 & 796 & 817 & 1{,}027 & 1{,}138 \\
    KDA & $C = 32$ & 880 & 905 & 948 & 990 & 1{,}085 \\
    Titans & $\dhid = 64$, $C = 16$ & 1{,}793 & 2{,}098 & 2{,}808 & 4{,}432 & 7{,}980 \\
    Titans & $\dhid = 64$, $C = 32$ & 1{,}202 & 1{,}215 & 1{,}641 & 2{,}399 & 4{,}097 \\
    \model & $\dhid = 64$, $C = 32$ (width-matched) & 990 & 1{,}023 & 1{,}233 & 1{,}303 & 1{,}444 \\
    \model & $\dhid = 256$, $C = 32$ (default) & 1{,}105 & 1{,}223 & 1{,}325 & 1{,}463 & 1{,}544 \\
    \midrule
    \multicolumn{2}{@{}l}{Default / width-matched \model} & 1.12 & 1.20 & 1.07 & 1.12 & 1.07 \\
    \multicolumn{2}{@{}l}{Default \model{} / KDA} & 1.25 & 1.35 & 1.40 & 1.48 & 1.42 \\
    \bottomrule
  \end{tabular}
\end{table}

\subsection{Ablation protocol}
\label{app:ablation}

All ablations use the 145M configuration of \tabref{tab:ladder_config} (12 layers, width
1024, 16 heads) with the same data, token budget (9.2B tokens), optimizer, learning-rate
schedule, and hardware configuration as the corresponding run of the ladder, and change one
component of the default \model configuration at a time: (i) the homogeneous-coordinate
bias of \secref{sec:practical}, removed; (ii) the hidden width $\dhid \in \{64, 128, 192,
256, 512\}$, i.e., one to four and eight times $\dk$; (iii) the activation $\sigma \in \{\text{identity},
\tanh, \text{SiLU}, \text{leaky ReLU}, \text{GELU}, \text{softmax}\}$; (iv) the error scale $\gamma
\in \{0, 0.001, 0.005, 0.01, 0.05, 0.1\}$ of \eqnref{eq:target_practical}, which bounds
each coordinate of the hidden correction and thus sets the effective step of the hidden
layer; and (v) the update rule of the hidden memory, the WY kernel (line~\ref{ln:wy} of
\algref{alg:toaster}) in place of the default apply-then-update step. Every row is compared with one run of the default configuration (C4 perplexity 15.687). The KDA reference is the 145M run of the scaling ladder
(\figref{fig:scaling}). Each run is evaluated once, at the end of training, on the C4
validation split at the 4,096-token context. The state per head is $(\dk+1)\dhid +
(\dhid+1)\dv$ with the bias (33,088 at the default width; 8,320, 16,576, 24,832, and 66,112 at
$\dhid = 64$, $128$, $192$, and $512$) and $\dk\dhid + \dhid\dv = 32{,}768$ without it. With the
identity activation the two memories compose to a single affine map whose linear part has rank at most $\dk$,
so that row isolates the effect of the nonlinearity from that of the larger state.

\paragraph{Discussion.}
The rows of \tabref{tab:ablation} separate
three factors in the 0.31 gain over KDA (single runs). The first is a larger state: KDA and \model have
nearly the same parameters (\appref{app:scaling_fit}), but \model holds about eight times
KDA's state, and perplexity improves monotonically as the hidden width grows from $\dk$ to
$4\dk$ (15.90 to 15.69, and to 15.62 at $8\dk$). A cascade with the identity activation, which has this state but,
like KDA, reads through a map whose linear part has rank at most $\dk$, recovers 0.20 of the gain; this
row also differs from KDA in the bias, the learned initial states, and the gate scales, so the 0.20
cannot be attributed to the cascade alone (\secref{sec:practical}, \appref{app:mixers}). The second is the nonlinearity, which adds a further 0.11 by letting the read use the wider state; GELU is the best of the six
activations tried, and a softmax, as in GSA, is the worst (18.19; \appref{app:additional}). Most of the 0.11 remains with $\gamma = 0$, where the hidden memory receives
no error signal and changes only through its forget gate, so that \model is a delta-rule
memory over a learned nonlinear feature map of the key~\citep{liu2026tttla}, and reaches
15.71. The hidden memory is not static in this row; with it frozen completely,
perplexity is 15.86 (\tabref{tab:frozen_hidden}). The third is target propagation, which lets the hidden memory learn in context as
well: it improves on $\gamma = 0$ by 0.02 with the default apply-then-update step. Advancing the hidden memory with the WY kernel instead, which lets each write see
the earlier ones, lowers perplexity by a further 0.05 (15.64, last row); the two changes together give 0.07 over $\gamma = 0$, but we did not run $\gamma = 0$ with the WY kernel, so the share of target propagation in that 0.07 is not isolated. The error scale is otherwise flat over $0.001$--$0.01$ and degrades only
at larger values, and removing the homogeneous-coordinate bias costs more than halving the
width. At 145M, most of the gain thus comes from a wide nonlinear memory that runs on
linear-memory kernels; learning in both layers adds a little to it, and advancing the hidden memory
exactly adds more, but the WY kernel cost $1.3$--$1.4\times$ the step time in preliminary runs,
which is why the cheaper rule is the default (\secref{sec:practical}).

\section{Full results}
\label{app:full_results}

\begin{table}[H]
  \caption{Validation loss (nats per token) and perplexity across the scaling ladder (lower
  is better; best per row within each block in bold), and the difference in loss between
  \model and KDA and between \model and Attention (negative favors \model). Single runs.}
  \label{tab:ladder_loss}
  \centering
  \small
  \setlength{\tabcolsep}{4.5pt}
  \begin{tabular}{@{}l ccc ccc cc@{}}
    \toprule
    & \multicolumn{3}{c}{Loss (nats)} & \multicolumn{3}{c}{Perplexity} & \multicolumn{2}{c}{$\Delta$ loss of \model vs.} \\
    \cmidrule(lr){2-4}\cmidrule(lr){5-7}\cmidrule(lr){8-9}
    Corpus / size & Attention & KDA & \model & Attention & KDA & \model & KDA & Attention \\
    \midrule
    \ladderlossrows
    \bottomrule
  \end{tabular}
\end{table}

\begin{figure}[H]
  \centering
  \includegraphics[width=\textwidth]{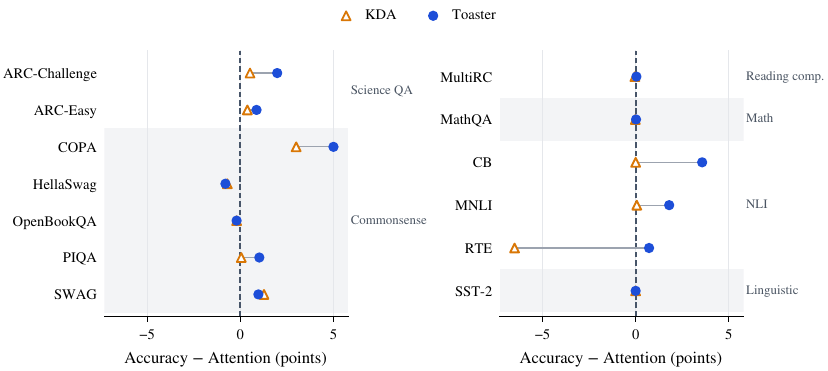}
  \caption{\textbf{Per-task three-shot accuracy relative to Attention at 1.2B}, for the
  three models trained along the ladder (points; right of the dashed line is better than
  Attention; single runs). \model is above Attention
  on 10 of 13 tasks and tied on SST-2; its largest margins (COPA, CB, RTE) fall on the
  smallest evaluation sets, and most other differences are within one binomial standard
  error (\appref{app:eval}).}
  \label{fig:downstream}
\end{figure}

\begin{figure}[H]
  \centering
  \includegraphics{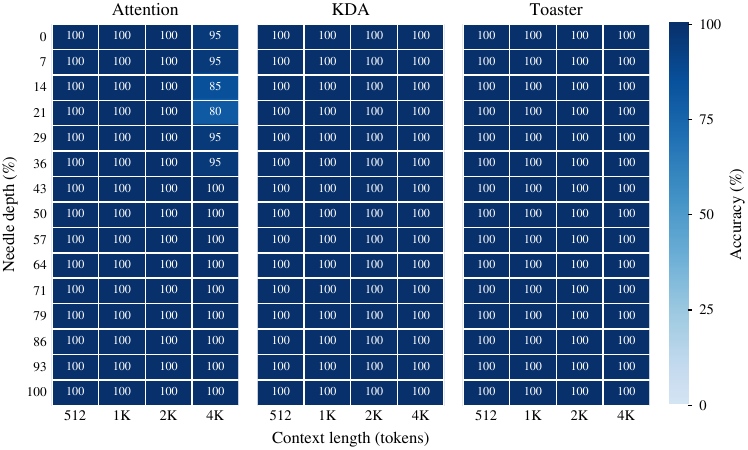}
  \caption{\textbf{Pressure test on S-NIAH-1 (passkey) at 1.2B.} Accuracy (\%) for every
  needle depth (0\%: needle at the start of the context, farthest from the query) and
  context length (\appref{app:niah}).}
  \label{fig:niah_s1}
\end{figure}

\begin{figure}[H]
  \centering
  \includegraphics{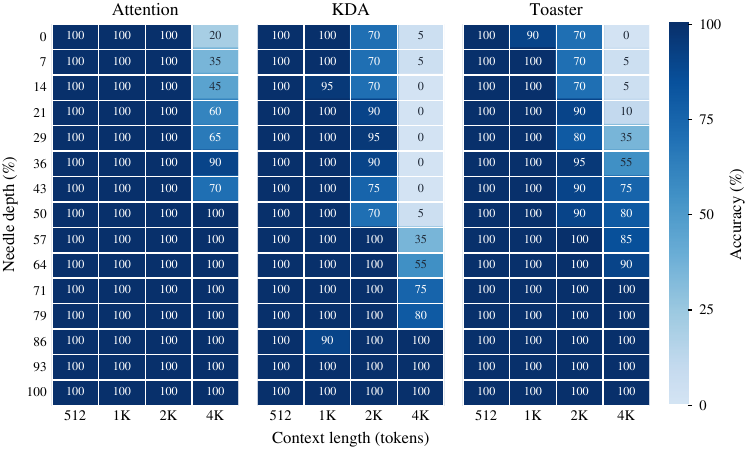}
  \caption{\textbf{Pressure test on S-NIAH-2 (number in essays) at 1.2B.} Same layout as
  \figref{fig:niah_s1}.}
  \label{fig:niah_s2}
\end{figure}

\begin{figure}[H]
  \centering
  \includegraphics{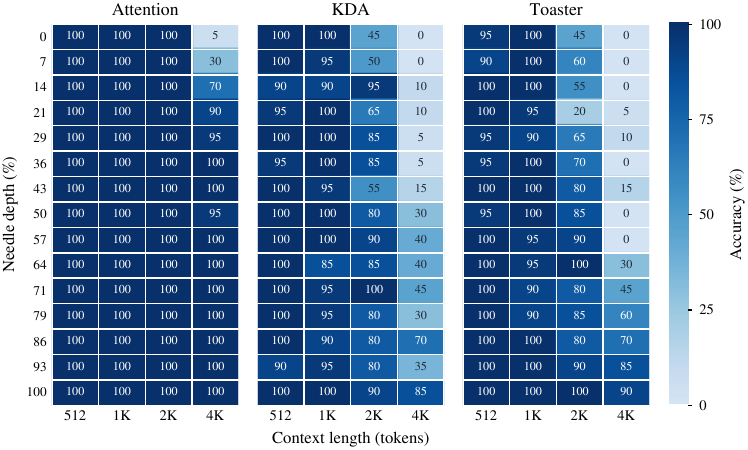}
  \caption{\textbf{Pressure test on S-NIAH-3 (UUID in essays) at 1.2B.} Same layout as
  \figref{fig:niah_s1}.}
  \label{fig:niah_s3}
\end{figure}

\paragraph{Pressure test.}
Figures~\ref{fig:niah_s1}--\ref{fig:niah_s3} show every cell of the test. Up to 1K tokens
every cell is at or near 100\% for all models and tasks, and the passkey
task (S-NIAH-1) stays at 100\% at every length for the recurrent models. Where the
haystack is real text, accuracy at 4K depends on the depth of the needle. On S-NIAH-2,
\model retrieves at least 75\% of the needles placed at depth $43\%$ or later and all of
those from depth $71\%$ on, whereas KDA reaches 75\% only from depth $71\%$ and 100\% from
$86\%$; on S-NIAH-3 both recurrent models fail for needles in the first half of the
context and recover only towards its end (\model at least 60\% from depth $79\%$, KDA above
60\% at two depths). Attention's errors are concentrated near the beginning of the context: it retrieves every needle from depth $50\%$ on in S-NIAH-2 and at least 90\% from depth $21\%$ on in S-NIAH-3, but only 20\% and 5\% of those at depth 0. At 2K the same ordering appears in milder form.

\begin{table}[H]
  \caption{Three-shot accuracy (\%) on all 13 downstream tasks at 1.2B (nominal), with all
  layers replaced by the sequence mixer (left) and in the 3:1 hybrid with global attention
  (right; \tabref{tab:downstream}), and the average within each skill category.
  {\setlength{\fboxsep}{1pt}\colorbox{bestcolor}{\textbf{Bold}}}: best per row within each
  group; {\setlength{\fboxsep}{1pt}\colorbox{secondcolor}{shaded}}: second best; rows on
  which a whole group ties (SST-2) are left unmarked.}
  \label{tab:downstream_full}
  \centering
  \footnotesize
  \setlength{\tabcolsep}{4.5pt}
  \begin{tabular}{@{}l cccc cccc@{}}
    \toprule
    & \multicolumn{4}{c}{All layers replaced} & \multicolumn{4}{c}{Hybrid (3 mixer : 1 attention)} \\
    \cmidrule(lr){2-5}\cmidrule(lr){6-9}
    Task & Attention & KDA & Titans & \model & SWA & KDA & Titans & \model \\
    \midrule
    \downstreamfullrows
    \bottomrule
  \end{tabular}
\end{table}

\begin{table}[H]
  \caption{Length extrapolation at 1.2B: mean loss (nats per token) of each model in
  4K-token windows of the 20K-token sequences of \figref{fig:extrapolation}, and the
  difference between \model and KDA (negative favors \model; the per-window standard error
  of the difference is about $0.001$ nats). The five 4K windows
  are consecutive and do not overlap; the last row pools 4--20K. The training context is 4K tokens.}
  \label{tab:extrapolation_windows}
  \centering
  \small
  \setlength{\tabcolsep}{5pt}
  \begin{tabular}{@{}l ccc c@{}}
    \toprule
    Window & Attention & KDA & \model & \model $-$ KDA \\
    \midrule
    0--4K   & 2.579 & 2.589 & 2.582 & $-$0.008 \\
    4--8K   & 2.673 & 2.493 & 2.479 & $-$0.014 \\
    8--12K  & 3.444 & 2.491 & 2.481 & $-$0.010 \\
    12--16K & 6.012 & 2.480 & 2.469 & $-$0.011 \\
    16--20K & 6.756 & 2.497 & 2.488 & $-$0.009 \\
    \midrule
    4--20K  & 4.721 & 2.490 & 2.479 & $-$0.011 \\
    \bottomrule
  \end{tabular}
\end{table}

\newcommand{\ablationrows}{%
KDA (linear memory) & 16.00 & \cellcolor{red!45}$+$0.31 \\
\model (default) & 15.69 & -- \\
\addlinespace[2pt]\multicolumn{3}{@{}l}{\textit{Bias}} \\
\quad w/o.\ homogeneous-coordinate bias & 15.86 & \cellcolor{red!31}$+$0.17 \\
\addlinespace[2pt]\multicolumn{3}{@{}l}{\textit{Hidden width}} \\
\quad $\dhid = \dk$ (64) & 15.90 & \cellcolor{red!39}$+$0.22 \\
\quad $\dhid = 2\dk$ (128) & 15.76 & \cellcolor{red!13}$+$0.07 \\
\quad $\dhid = 3\dk$ (192) & 15.72 & \cellcolor{red!7}$+$0.04 \\
\quad $\dhid = 8\dk$ (512) & 15.62 & \cellcolor{blue!13}$-$0.07 \\
\addlinespace[2pt]\multicolumn{3}{@{}l}{\textit{Activation $\sigma$}} \\
\quad identity (affine cascade) & 15.80 & \cellcolor{red!20}$+$0.11 \\
\quad tanh & 15.82 & \cellcolor{red!24}$+$0.13 \\
\quad SiLU & 15.79 & \cellcolor{red!18}$+$0.10 \\
\quad leaky ReLU & 15.71 & \cellcolor{red!4}$+$0.02 \\
\quad softmax & 18.19 & \cellcolor{red!45}$+$2.50 \\
\addlinespace[2pt]\multicolumn{3}{@{}l}{\textit{Error scale $\gamma$ (hidden-layer step)}} \\
\quad $\gamma = 0$ (no hidden-layer error) & 15.71 & \cellcolor{red!3}$+$0.02 \\
\quad $\gamma = 0.001$ & 15.69 & 0.00 \\
\quad $\gamma = 0.005$ & 15.68 & \cellcolor{blue!1}$-$0.01 \\
\quad $\gamma = 0.05$ & 15.70 & \cellcolor{red!3}$+$0.02 \\
\quad $\gamma = 0.1$ & 15.74 & \cellcolor{red!9}$+$0.05 \\
\addlinespace[2pt]\multicolumn{3}{@{}l}{\textit{Hidden-memory update}} \\
\quad WY kernel instead of apply-then-update & 15.64 & \cellcolor{blue!9}$-$0.05 \\}

\begin{table}[H]
  \caption{\textbf{Ablation study} at 145M (9.2B tokens; \appref{app:ablation}). Each row changes one
  component of the default \model (bias, $\dhid = 4\dk$, GELU, $\gamma = 0.01$, apply-then-update) and
  reports C4 perplexity and its difference $\Delta$ to the default (red: worse, blue: better). The
  default row is the ablation's own run (15.687; the ladder's 145M run gives 15.690).}
  \label{tab:ablation}
  \centering
  \small
  \setlength{\tabcolsep}{7pt}
  \begin{tabular}{@{}l cc@{}}
    \toprule
    \model ablations (145M) & C4 ppl ($\downarrow$) & $\Delta$ \\
    \midrule
    \ablationrows
    \bottomrule
  \end{tabular}
\end{table}

\section{Additional ablations and diagnostics}
\label{app:additional}

This section collects ablations and diagnostics that were added after the first version of
this paper. Unless stated otherwise, they use the 145M configuration and the protocol of
\appref{app:ablation}: one run each, evaluated on the C4 validation split at the 4,096-token
context.

\paragraph{Freezing the hidden memory.}
With $\gamma = 0$ (\tabref{tab:ablation}), the hidden memory $\Sone$ receives no error signal,
but it is not static: its input-dependent forget gate still changes it at every chunk. To
separate a fixed nonlinear feature map from the adaptation of $\Sone$ in context, we trained a
145M \model in which $\Sone$ is frozen completely at its learned initial state
(\tabref{tab:frozen_hidden}). With the same architecture and state size, it reaches 15.86: a fixed
feature map gives 0.14 of the 0.31 gain over KDA, and adapting $\Sone$ in context gives the
remaining 0.17, reaching 15.71 through the forget gate alone and 15.69 with the error signal.
Over a 4K context, $\Sone$ moves from its initial state by 4.7\% in relative Frobenius norm at
145M and by 3.4\% at 1.2B, and by 31.5\% in the model of \tabref{tab:chunk_size} (top), where the default kernel advances both layers. In that model,
freezing $\Sone$ at inference alone raises perplexity by 0.39 on C4 (15.77 to 16.16) and by 0.94
on PG19 (22.33 to 23.28).

\begin{table}[H]
  \caption{\textbf{Freezing the hidden memory} (145M, C4 perplexity). All \model rows share the
  architecture and the state size; the $\gamma = 0$ row is that of \tabref{tab:ablation}. The
  middle column gives the relative change of $\Sone$ over a 4K context.}
  \label{tab:frozen_hidden}
  \centering
  \small
  \begin{tabular}{@{}l cc@{}}
    \toprule
    Model (145M) & $\Sone$ changes & C4 ppl ($\downarrow$) \\
    \midrule
    KDA & -- & 16.00 \\
    \model, $\Sone$ frozen completely & no (0\%) & 15.86 \\
    \model, $\gamma = 0$ (forget gate still active) & yes & 15.71 \\
    \model, default (full target propagation) & yes (4.7\%) & 15.69 \\
    \bottomrule
  \end{tabular}
\end{table}

\paragraph{Chunk size.}
The chunkwise algorithm evaluates the cross-layer signals once per chunk, at the chunk-start
states, and is exact only for $C = 1$ (\secref{sec:chunk}). We measured the effect of this
approximation on a 145M model with hidden width $\dhid = \dk = 64$, a quarter of the default, and the default kernel on both layers. This kernel advances each memory token by token, with apply-then-update on the hidden memory and update-then-apply on the output memory. The model is a separate run, and its configuration is not a row of \tabref{tab:ablation}: the $\dhid = \dk$ row (15.90) updates the hidden memory once per chunk, and the WY-kernel row runs the WY kernel on both layers at the default width. For this model, decoding with chunk size
$C' = 1$ is the exact coupled per-token recurrence. Evaluated with
$C' \in \{1, 4, 8, 16, 32, 64\}$, the model trained with $C = 32$ moves smoothly from 15.97 at
$C' = 1$ to 15.77 at $C' = 32$ and 15.79 at $C' = 64$ (\tabref{tab:chunk_size}, top), and the
hidden-memory states of its first block differ from those at $C' = 1$ by 1.7\% in relative
Frobenius norm after 4K tokens. We also trained the default \model with chunk sizes
$C \in \{8, 16, 32, 64, 128, 256\}$ (\tabref{tab:chunk_size}, bottom). Smaller chunks refresh the
cross-layer signals more often and reach 15.59 ($C = 8$) and 15.56 ($C = 16$), against 15.69 at
the default $C = 32$; $C \in [32, 128]$ stays at 15.67--15.69, and $C = 256$ reaches 15.74, still
below KDA's 16.00.

\begin{table}[H]
  \caption{\textbf{Effect of the chunk size} (145M, C4 perplexity). Top: one model with hidden width $\dhid = \dk$ and the default kernel on both layers (token by token, with apply-then-update on the hidden memory and update-then-apply on the output memory; a separate run, not a row of \tabref{tab:ablation}), trained with $C = 32$ and evaluated with decoding chunk size $C'$; $C' = 1$ is
  the exact per-token recurrence. Bottom: the default \model ($\dhid = 4\dk$) trained with chunk size $C$.}
  \label{tab:chunk_size}
  \centering
  \small
  \setlength{\tabcolsep}{6pt}
  \begin{tabular}{@{}l cccccc@{}}
    \toprule
    Decoding chunk size $C'$ & 1 & 4 & 8 & 16 & 32 & 64 \\
    \midrule
    Default kernel on both layers & 15.97 & 15.90 & 15.84 & 15.79 & 15.77 & 15.79 \\
    \midrule
    Training chunk size $C$ & 8 & 16 & 32 & 64 & 128 & 256 \\
    \midrule
    \model (default) & 15.59 & 15.56 & 15.69 & 15.67 & 15.69 & 15.74 \\
    \bottomrule
  \end{tabular}
\end{table}

\paragraph{KDA with a larger state.}
KDA's state is tied to its head dimension, so it can be enlarged only by trading heads for
wider heads. At a fixed backbone ($H \cdot \dk = D = 1024$, which leaves the sizes of the query, key, value, gate, and output projections unchanged), 8, 4, and 2 heads of dimension 128, 256, and 512 give 2, 4, and 8 times
the state of the default KDA and lower C4 perplexity from 16.00 to 15.96, 15.84, and 15.79
(\tabref{tab:kda_state}). This closes about two-thirds of the gap to \model, which stays
0.06--0.10 below KDA at every matched state size while keeping its 16 heads of dimension 64.

\begin{table}[H]
  \caption{\textbf{KDA with a larger state} (145M, C4 perplexity). KDA trades heads for wider
  heads at fixed $H \cdot \dk = D = 1024$; \model keeps 16 heads of dimension 64 and widens its
  hidden layer.}
  \label{tab:kda_state}
  \centering
  \small
  \begin{tabular}{@{}l cccc@{}}
    \toprule
    Total state (relative to default KDA) & $1\times$ & $2\times$ & $4\times$ & $8\times$ \\
    \midrule
    KDA, heads $\times$ head dimension & $16 \times 64$ & $8 \times 128$ & $4 \times 256$ & $2 \times 512$ \\
    KDA, C4 ppl & 16.00 & 15.96 & 15.84 & 15.79 \\
    \midrule
    \model, hidden width $\dhid$ & -- & 64 & 128 & 256 \\
    \model, C4 ppl & -- & 15.90 & 15.76 & 15.69 \\
    \bottomrule
  \end{tabular}
\end{table}

\paragraph{Wider hidden layer and softmax activation.}
\tabref{tab:ablation} also covers a wider hidden layer and a softmax activation. Doubling the hidden width to $\dhid = 512$
lowers perplexity from 15.69 to 15.62, so perplexity improves steadily with the width. A softmax
between the two memories, as in GSA~\citep{zhang2024gsa}, gives 18.19, 2.50 above GELU; for this
row, the hidden target of \eqnref{eq:target_practical} uses the Jacobian of the softmax in place
of the pointwise derivative $\sigma'$. Our reading is that a softmax restricts the hidden
activations to a distribution over the hidden units, which suits slot selection but leaves the
output memory less room than an unconstrained activation.

\paragraph{Retrieval in the hybrids.}
We ran the needle-in-a-haystack tests of \tabref{tab:niah}, with the same protocol, on the
four 1.2B hybrids of \tabref{tab:downstream} (\tabref{tab:hybrid_niah}). At 4K, the \model
hybrid scores 99.7 / 89.3 / 77.3\% on S-NIAH-1 / 2 / 3 and is the best of the four hybrids on all
three tasks. One global attention layer in four lifts its accuracy on S-NIAH-3 at 4K from 27.3\%
with all layers replaced (\tabref{tab:niah}) to 77.3\%.

\begin{table}[H]
  \caption{\textbf{Single needle-in-a-haystack accuracy (\%) of the 3:1 hybrids at 1.2B}, averaged
  over 15 needle positions, by context length (protocol of \tabref{tab:niah}).}
  \label{tab:hybrid_niah}
  \centering
  \small
  \setlength{\tabcolsep}{4.2pt}
  \begin{tabular}{@{}l cccc cccc cccc@{}}
    \toprule
    & \multicolumn{4}{c}{S-NIAH-1 (passkey)} & \multicolumn{4}{c}{S-NIAH-2 (number)} & \multicolumn{4}{c}{S-NIAH-3 (UUID)} \\
    \cmidrule(lr){2-5}\cmidrule(lr){6-9}\cmidrule(lr){10-13}
    Hybrid & 512 & 1K & 2K & 4K & 512 & 1K & 2K & 4K & 512 & 1K & 2K & 4K \\
    \midrule
    SWA    & 100.0 & 100.0 & 100.0 & 89.3 & 99.3 & 100.0 & 100.0 & 78.7 & 98.7 & 93.7 & 91.7 & 56.3 \\
    KDA    & 100.0 & 99.7 & 99.3 & 98.7 & 100.0 & 99.7 & 100.0 & 73.3 & 93.0 & 96.7 & 81.7 & 44.0 \\
    Titans & 100.0 & 100.0 & 100.0 & 84.7 & 100.0 & 99.7 & 93.7 & 81.7 & 99.3 & 91.0 & 96.0 & 59.7 \\
    \model & 100.0 & 100.0 & 100.0 & 99.7 & 100.0 & 100.0 & 97.3 & 89.3 & 99.7 & 90.7 & 96.3 & 77.3 \\
    \bottomrule
  \end{tabular}
\end{table}

\end{document}